\documentclass[10pt,twocolumn,letterpaper]{article}

\usepackage{iccv}
\usepackage{times}
\usepackage{epsfig}
\usepackage{graphicx}
\usepackage{amsmath}
\usepackage{amssymb}
\usepackage{url}

\newcommand{\nameofmethod}{GETAvatar}
\usepackage{float}
\usepackage{graphicx}
\usepackage{color}
\usepackage{xcolor}
\usepackage{multirow}
\usepackage{makecell}
\usepackage{subfig}
\usepackage{colortbl}
\usepackage{utfsym}
\usepackage{bbding}
\usepackage{pifont}
\usepackage{booktabs}

\usepackage[breaklinks=true,bookmarks=false]{hyperref}

\iccvfinalcopy % *** Uncomment this line for the final submission

\def\iccvPaperID{****} % *** Enter the ICCV Paper ID here

\ificcvfinal\fi

\begin{document}

%%%%%%%%% TITLE
\title{\nameofmethod{}: Generative Textured Meshes for Animatable Human Avatars}

\vspace{-5mm}
\author{Xuanmeng Zhang$^{1,2*}$ 
\quad Jianfeng Zhang$^{2,3*}$  
\quad Rohan Chacko$^{2}$  \\
Hongyi Xu$^{2}$  \quad 
Guoxian Song$^{2}$ \quad 
Yi Yang$^{4}$ \quad 
Jiashi Feng$^{2}$ \\
$^1${ReLER, AAII, University of Technology Sydney} 
\quad $^2${ByteDance} \\
\quad $^3$ National University of Singapore 
\quad $^4${ReLER, CCAI, Zhejiang University} \\
% {\tt\small \{zhangxuanmeng.zxm, zdzheng12\}@gmail.com} \\
% {\tt\small \{daiheng.gdh, zhangbang.zb, panpan.pp\}@alibaba-inc.com yangyics@zju.edu.cn}
}

% \maketitle

\twocolumn[{%
\renewcommand\twocolumn[1][]{#1}%
\maketitle
\ificcvfinal\thispagestyle{empty}\fi
\captionsetup{type=figure}
\vspace{-13mm}
\begin{center}
\centering
\includegraphics[width=1.0 \linewidth]{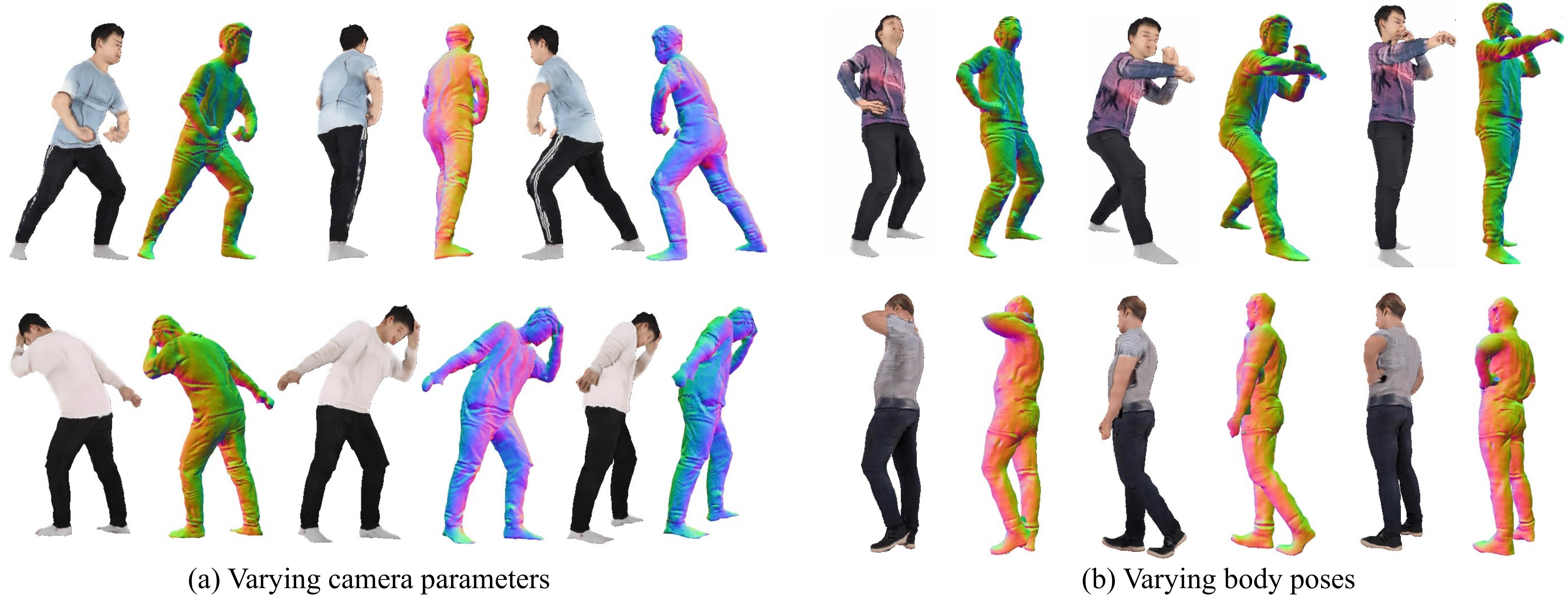}
\vspace{-7mm}
 \captionof{figure}{\nameofmethod{} generates controllable human avatars with diverse textures and detailed geometries under full control over camera poses and body poses.  Please refer to the Appendix for more multi-view and animation results.}
\label{fig:teaser}
\vspace{-3mm}
\end{center}%
}]
%-------------------------------------------------------------------------

\let\thefootnote\relax\footnotetext{*Equal contribution.}

%%%%%%%%% ABSTRACT
\begin{abstract}
We study the problem of 3D-aware full-body human generation, aiming at creating animatable human avatars with high-quality textures and geometries. 
Generally, two challenges remain in this field: 
i) existing methods struggle to generate geometries with rich realistic details such as the wrinkles of garments;
ii) they typically utilize volumetric radiance fields and neural renderers in the synthesis process, making high-resolution rendering non-trivial. 
To overcome these problems, we propose \textbf{\nameofmethod{}}, a \textbf{G}enerative model that directly generates \textbf{E}xplicit \textbf{T}extured 3D meshes for animatable human \textbf{Avatar}, with photo-realistic appearance and fine geometric details.
Specifically, we first design an articulated 3D human representation with explicit surface modeling, 
% \textcolor{red}{We enrich the generated humans with realistic surface details by conducting adversarial training of 3D normal field with the 3D scan data.}
and enrich the generated humans with realistic surface details by learning from the 2D normal maps of 3D scan data.
Second, with the explicit mesh representation, we can use a  rasterization-based renderer to perform surface rendering, allowing us to achieve high-resolution image generation efficiently.
Extensive experiments demonstrate that \nameofmethod{} achieves state-of-the-art performance on 3D-aware human generation both in appearance and geometry quality. 
Notably, GETAvatar can generate images at $512^2$ resolution with 17FPS and  $1024^2$ resolution with 14FPS, improving upon previous methods by $2\times$. Our code and models will be at \url{https://getavatar.github.io/}.
\end{abstract}

%%%%%%%%% BODY TEXT
\section{Introduction}
Generating  high-quality 3D human avatars with explicit control over  camera poses, body poses and shapes has been a long-standing challenge in computer vision and graphics. It has wide applications in video games, AR/VR, and movie production.
Recently, 3D-aware generative models have demonstrated impressive results in producing multi-view-consistent images of 3D shapes~\cite{schwarz2020graf,chan2021pi,Niemeyer2020GIRAFFE,chan2022efficient,or2022stylesdf,gao2022get3d}.
However, despite their success in modeling relatively simple and  rigid objects, 
it remains challenging for modeling dynamic human bodies with large articulated motions.
The main reason is that these 3D GANs are not designed to handle body deformations, such as variations in human shapes and poses. 
Thus, they struggle to manipulate or animate the generated avatars given the control signals.
% according to the desired instructions.

Some recent works~\cite{noguchi2022unsupervised,bergman2022generative,hong2022eva3d,zhang2022avatargen}  have incorporated human priors~\cite{loper2015smpl} into 3D-aware generative models~\cite{chan2021pi,chan2022efficient} to generate animatable 3D human avatars.
However, these methods face two challenges.
First, the generated human avatars lack fine geometric details, such as cloth wrinkles and hair, which are highly desirable  for the photo-realistic 3D human generation. 
Second, existing methods~\cite{noguchi2022unsupervised,bergman2022generative,hong2022eva3d,zhang2022avatargen} adopt volumetric neural renderers in the synthesis process, which suffers from high computational
costs, making high-resolution rendering non-trivial.

In this work, we propose \nameofmethod{}, a generative model that produces explicit textured 3D meshes with rich surface details (see Fig.~\ref{fig:teaser})  for {animatable} human avatars.
Previous methods~\cite{noguchi2022unsupervised,bergman2022generative,hong2022eva3d,zhang2022avatargen} model the human body with implicit geometry representations, \ie,
density fields and signed distance fields, which produce either noisy or over-smoothed geometries due to the lack of explicit surface modeling and insufficient geometric supervision.
To improve the geometry quality of the generated humans, different from previous methods, we propose to model fine geometric details with a normal field~\cite{gropp2020implicit}, which associates a normal vector with each point on the  surface.
The direction and magnitude of the normal vector provide crucial geometric information to represent the detailed human body surface.

Specifically, we design a body-controllable articulated 3D human representation with body deformation modeling and explicit surface modeling.
The former allows us to deform the generated humans to target pose and shape, and the latter enables us to extract the underlying human body surface as an explicit mesh in a differentiable manner.
Based on the extracted meshes, we further construct a normal field to depict the detailed surface of the generated humans. 
The normal field enables the model to capture realistic geometric details from 2D normal maps that are rendered from available 3D human scans, improving  geometry quality and resulting in higher fidelity appearance generation significantly.
Besides, existing methods~\cite{noguchi2022unsupervised,bergman2022generative,hong2022eva3d} struggle to render high-resolution images as the volume rendering process require intensive memory and computational costs.
The main issue is that volumetric radiance field and neural renders perform volume sampling on both occupied and free regions~\cite{mildenhall2020nerf} that do not contribute to the rendered images, resulting in computational inefficiency.
In contrast, \nameofmethod{} benefits from the proposed explicit representation and thus can generate textured meshes in a differentiable manner. 
With the extracted mesh surface, we can render high-resolution images up to $1024^2$ with a highly efficient rasterization-based surface renderer~\cite{laine2020modular}.

To validate the effectiveness of \nameofmethod{}, we conduct extensive experiments on two 3D human datasets~\cite{renderpeople,tao2021function4d}.
The quantitative and qualitative  results demonstrate that \nameofmethod{} consistently outperforms previous methods in terms of both visual and geometry quality (see Fig.~\ref{fig:sota}).
Overall, our work makes the following contributions:
\begin{enumerate}
\item  
We propose a generative model, \nameofmethod{}, that enables high-quality 3D-aware human generation, with full control over camera poses, body shapes, and human poses.
\item
We propose to model the complex body surface using a 3D normal field, which significantly improves the geometric details of the generated clothed humans.

\item
 We design an articulated 3D human representation with differentiable surface modeling.
The explicit mesh representation supports $360^{\circ}$ free-view, high-resolution image synthesis  ($1024^2$) 
for the generated avatars, and supports normal map rendering.

\item  
Our \nameofmethod{} can be applied to a wide range of tasks, such as re-texturing, single-view 3D reconstruction, and re-animation.

\end{enumerate}

%------------------------------------------------------------------------
\section{Related Work}
\noindent \textbf{3D-aware Generative Models.}
In recent years, 3D-aware image generation~\cite{schwarz2020graf} has  gained a surge of interest.
To generate objects and scenes in 3D space, 3D-aware generative models~\cite{chan2021pi,Niemeyer2020GIRAFFE,chan2022efficient,gao2022get3d,or2022stylesdf,zhang2022multi,xu2022pv3d}  incorporate 3D representations into generative adversarial networks, such as point clouds, 3D primitives~\cite{liao2020towards}, voxels~\cite{nguyen2019hologan}, meshes~\cite{gao2022get3d}, and neural radiances fields~\cite{schwarz2020graf}.
Among them, NeRF-based generative models~\cite{chan2021pi,Niemeyer2020GIRAFFE,chan2022efficient,or2022stylesdf,zhang2022multi} have become the dominating direction of 3D generation due to the high-fidelity image synthesis and 3D consistency.
EG3D~\cite{chan2022efficient} introduces an efficient explicit-implicit framework with
a triplane hybrid representation.
% To obtain detailed 3D surfaces, 
StyleSDF~\cite{or2022stylesdf} combines an SDF-based 3D volume renderer and a style-based 2D generator to obtain 3D surfaces.
These methods typically perform volume rendering at a low resolution and then adopt a super-resolution module as the 2D decoder to get high-resolution results.
Recently, Gao~\etal~\cite{gao2022get3d} propose GET3D for synthesizing textured meshes for static rigid objects.
In this work, we take inspiration from GET3D~\cite{gao2022get3d} and propose a generative model for animatable human avatars.

\noindent \textbf{3D Human Generation.}
Recently, some works~\cite{noguchi2022unsupervised,bergman2022generative,hong2022eva3d,zhang2022avatargen} tackle the 3D human generation by combing 3D GANs with human representations.
Noguchi~\etal introduce ENARF~\cite{noguchi2022unsupervised} to learn articulated geometry-aware representations from 2D images.
Bergman~\etal propose Generative Neural Articulated Radiance Fields (GNARF)~\cite{bergman2022generative} to implement the generation and animation of human bodies.
Built on EG3D~\cite{chan2022efficient}, AvatarGen~\cite{zhang2022avatargen} adopts signed distance fields (SDFs) as geometry proxy to synthesize clothed 3D human avatars.
By dividing the human body into local parts, Hong~\etal propose a compositional NeRF representation for 3D human generation~\cite{hong2022eva3d}.
However, these methods fail to synthesize high-resolution images, and also cannot generate intricate geometric details of garments.
In contrast, \nameofmethod{} exploits an explicit mesh representation and a 3D normal field for human geometry modeling, thus is capable of generating 
human avatars with  realistic  details and achieves high-resolution photo-realistic image rendering.

\section{Preliminaries} \label{sec:preliminaries}
Our method involves triplane representation~\cite{chan2022efficient} and SMPL human model~\cite{loper2015smpl}.
 Here we provide a brief introduction to them. More details can be found in the original papers~\cite{chan2022efficient,loper2015smpl}.

\noindent \textbf{Triplane 3D Representation.}
Recently, EG3D~\cite{chan2022efficient} proposes an expressive and efficient 3D representation named triplanes for 3D generation.
Triplane contains three orthogonal axis-aligned feature planes with a shape of  $N \times N \times C$, where $N$ and $C$ denote the spatial resolution and the number of channels respectively.
Given any 3D points $\textbf{x}\in \mathbb{R}^{3}$, its feature can be extracted by projection and bi-linear lookups on triplanes.
Then, we can decode the aggregated triplane features into neural fields, \ie, color and signed distance.

\noindent \textbf{SMPL.}
Skinned Multi-Person Linear model (SMPL)~\cite{loper2015smpl} is a parametric human model that  represents a wide range of human body poses and shapes.
It defines a parameterized deformable mesh $\mathcal{M}(\beta, \theta)$, where a template mesh is deformed by linear blend skinning~\cite{lewis2000pose} with $\theta$  and $\beta$ representing articulated pose and shape parameters. 
It provides an articulated geometric proxy to the underlying dynamic human body.

\section{Method}

Our goal is to generate animatable human avatars  with full control over their camera views,  body poses and shapes.
To achieve this, we propose \nameofmethod{} for explicit textured 3D human meshes generation with  high-quality appearance and rich geometric details (\eg, clothing wrinkles and hairs).
Different from previous 3D human generation methods~\cite{noguchi2022unsupervised,bergman2022generative,zhang2022avatargen,hong2022eva3d},
\nameofmethod{} adopts an explicit ariticulated 3D representation and thus supports $360^{\circ}$ free-view, high-resolution ($1024^2$) and normal map rendering.

%-------------------------------------------------------------------------
\begin{figure*}[t]
\centering
\vspace{-10mm}
  \includegraphics[width=1\linewidth]{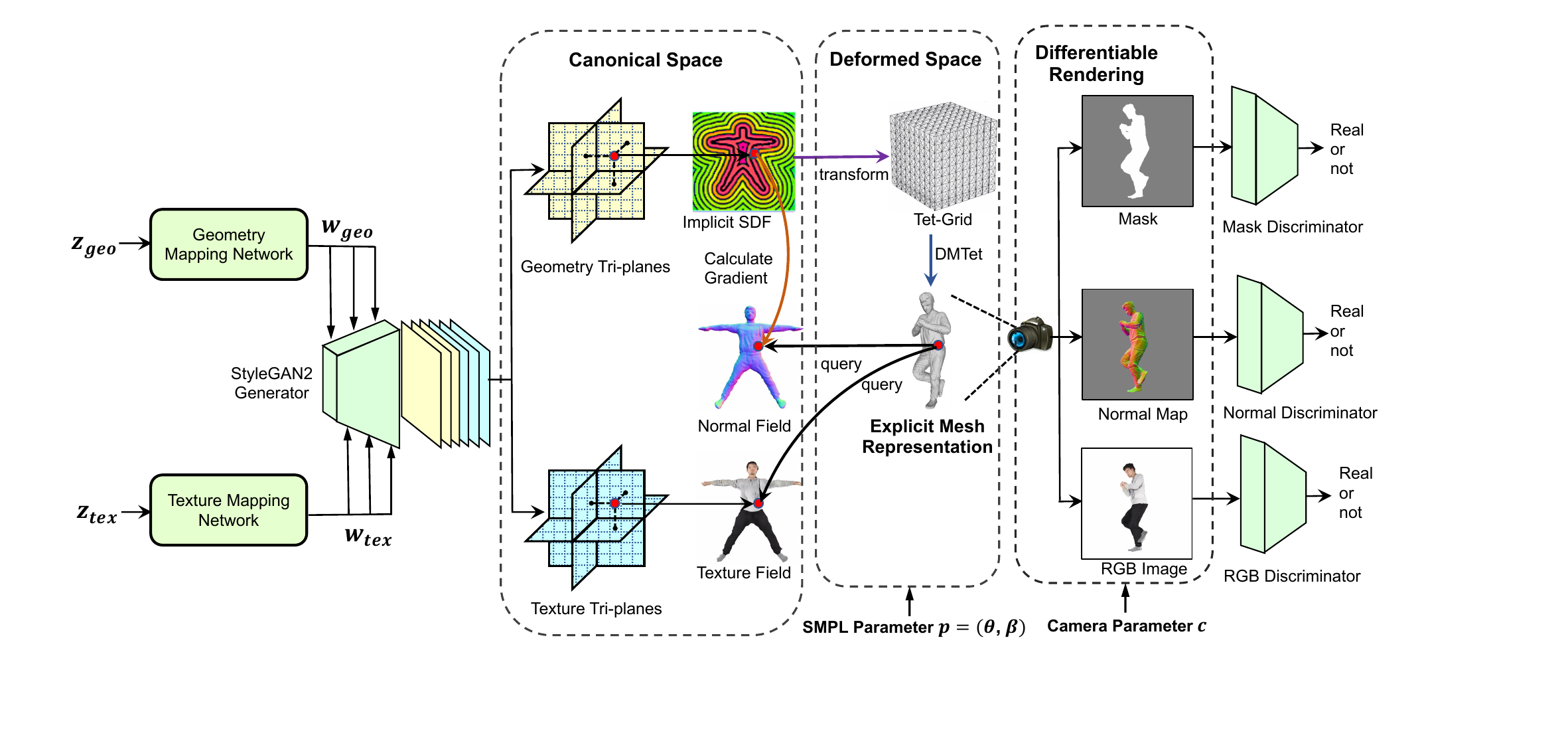}
  \caption{\textbf{The pipeline of \nameofmethod{}.}
\textbf{\uppercase\expandafter{\romannumeral1}. Generator.}
Given latent codes $z_{geo}$ and $z_{tex}$ sampled from Gaussian distribution, \nameofmethod{} generates the geometry triplanes $T_{geo}$ and the texture triplanes $T_{tex}$ via the StyleGAN2 generator backbone.
\textbf{\uppercase\expandafter{\romannumeral2}. Canonical Space.}
We model the canonical human representation with the signed distance, normal, and texture fields.  
\textbf{\uppercase\expandafter{\romannumeral3}. Deformed Space.}
To generate the target human avatar defined by SMPL~\cite{loper2015smpl} parameter $p$, we use a SMPL-guided deformation to transform the canonical signed distance field into a deformable tetrahedral grid in the deformed space, and then employ the DMTet~\cite{shen2021deep} to extract the underlying 3D mesh.
For every point at the surface of the generated mesh, the color and the normal can be obtained by querying the normal field and texture field from the corresponding location in the canonical space.
\textbf{\uppercase\expandafter{\romannumeral4}. Differentiable Rendering.}
To achieve the high-resolution image rendering, we adopt a differentiable rasterizer to render the 3D mesh into an RGB image, a normal map, and a mask from camera pose $c$.
\textbf{\uppercase\expandafter{\romannumeral5}. Adversarial Training.}
We use three discriminators~\cite{karras2020analyzing} to classify whether the input RGB image, normal map, and  mask are real or not.
}
\vspace{-5mm}
\label{fig:framework}
\end{figure*}
%-------------------------------------------------------------------------

\subsection{Overview}

\noindent \textbf{Framework.}
% \noindent \textbf{Problem definition.}
Given two latent codes $z_{geo}$ and $z_{tex}$ randomly sampled from  Gaussian distribution, a camera parameter $c$ consists of camera intrinsics and extrinsics, a SMPL~\cite{loper2015smpl} parameter $p = (\theta, \beta)$ that includes the human pose  $\theta$  and shape  $\beta$ parameters, {\nameofmethod{} first generates a human avatar mesh with the specified body attributes $p$}, and then synthesizes the corresponding RGB image, normal map, and foreground mask from the view defined by camera $c$. 
Here we define the 3D space corresponding to the target human representation with SMPL parameter $p$ as the \textit{deformed space}, and a \textit{canonical space} with a body pose- and shape-independent template human representation. 

% \noindent \textbf{Framework.}
We formulate the animatable human generation as a ``\textit{body deformation and explicit surface modeling}” process.
 The core idea is to first deform the implicit canonical human representation (triplane-based signed distance field) 
 to the target pose and shape via body deformation, and then we model the body surface with an explicit mesh representation and a normal field in the deformed space.
The overview of the proposed framework is shown in Fig.~\ref{fig:framework}.
Specifically, we first generate a shape- and pose-independent implicit human representation via two triplane branches~\cite{chan2022efficient}  in the canonical space.
To generate  human avatar with the desired body shape and pose, we deform the implicit human representation from the canonical space to the deformed space with the guidance of the SMPL model~(Sec.~\ref{sec:body_deformation_modeling}).
To model the body surface with fine details, 
we extract
the explicit 3D human body mesh from the signed distance fields in a differentiable manner~\cite{shen2021deep}, and improve the details of human body surface with a normal field (Sec.~\ref{sec:explicit_surface_modeling}).
After that, we render the generated human mesh into a 2D mask, normal map, and RGB image via an efficient differentiable surface renderer~\cite{laine2020modular} (Sec.~\ref{sec:rendering}), and train the whole framework via adversarial training~\cite{karras2020analyzing}~(Sec.\ref{sec:adv_training}).

\subsection{Controllable 3D Human Modeling} \label{sec:body_deformation_modeling}

\noindent \textbf{Canonical Human Generation.}
\nameofmethod{} is built upon 3D GANs~\cite{chan2022efficient,gao2022get3d}.
We model the geometry and texture of the canonical human with two separate triplane branches~\cite{chan2022efficient}, allowing for the disentanglement of geometry and appearance~(see Fig.~\ref{fig:framework}).
Given two latent codes $z_{geo}$ and $z_{tex}$ sampled from   Gaussian distribution, the geometry and texture mapping networks produce two intermediate latents  $w_{geo}$ and $w_{tex}$ to control the generation of the geometry and  texture triplanes.
For the geometry branch,  we model the canonical human representation as a signed distance field (SDF).
Specifically, given any 3D points in the canonical space, we query its feature from the geometry triplane, and adopt an MLP conditioned by $w_{geo}$  to decode the queried feature as the signed distance value.
However, directly modeling the signed distance field of clothed humans is challenging due to complex pose and shape variations. 
Therefore, instead of directly predicting the signed distance value, we predict a signed distance offset from the surface of the SMPL template~\cite{yifan2021geometry}.
For the texture branch, similar to the geometry branch, we use an MLP  to map the queried texture triplane features to color value. In this process, we condition the MLP on both $w_{geo}$ and $w_{tex}$, as the texture generation can also be influenced by changes in geometry.

\noindent \textbf{SMPL-guided Deformation.}
To warp the generated canonical human to a desired pose $\theta$ and shape $\beta$, we establish a correspondence mapping between the canonical space and the deformed space. For any point $\mathbf{x_d}$ in the deformed space ($\theta$, $\beta$), we aim to find its corresponding point $\mathbf{x_c}$ in the canonical space via the body deformation process.
It is intuitive to exploit the 3D human model SMPL~\cite{loper2015smpl} as deformation guidance.
Specifically, we generalize the linear blend skinning process~\cite{lewis2000pose} of the SMPL model from the coarse naked body to our generated clothed human. 
The core idea  is to associate each  point with its closest vertex on the deformed SMPL mesh  $\mathcal{M}( \theta,\beta)$,  assuming they undergo the same kinematic changes between the deformed and canonical spaces. Specifically, for a point $\mathbf{x_d}$ in the deformed space, we first find its nearest vertex $v^*$ in the SMPL mesh.
Then we use the skinning weights of $v^*$ to un-warp $\mathbf{x_d}$ to  $\mathbf{x_c}$ in the canonical space:
\begin{equation}
\begin{aligned}
\label{eq:warp}
\mathbf{x_c} &= \left(\sum\limits_{i=1}^{N_j} s_i^* \cdot B_i (\theta, \beta) \right)^{-1} \cdot \mathbf{x_d}, 
\end{aligned}
\end{equation}
where $N_j = 24 $ is the number of joints, $s_i^*$ is the skinning weight of vertex $v^*$  \wrt the $i$-th joint, 
$B_i (\theta, \beta)$ is the bone transformation matrix of join $i$.
With the SMPL-guided body deformation process, we can deform the canonical human to any desired pose and shape, enabling controllable human generation.

\subsection{Explicit Surface Modeling} \label{sec:explicit_surface_modeling}
Although the above pipeline can achieve controllable human generation, the resulting geometry produced by the implicit SDF is often noisy or over-smoothed (See 2nd row of Fig.~\ref{fig:ablation}), 
due to the lack of powerful geometry representation and insufficient geometric supervision.
To resolve this issue, we propose utilizing an \textbf{explicit mesh representation} for the human geometry modeling, and a \textbf{normal filed} build upon this explicit representation for generating the fine geometric details, \eg, hair, face, and cloth wrinkles.

To achieve explicit surface modeling  in the deformed space, we extract the mesh of the generated human avatars under the desired poses and shapes through a differentiable surface modeling technique, \ie, Deep Marching Tetrahedra (DMTet)~\cite{shen2021deep}. 
Specifically, DMTet represents the surface of humans with a discrete signed distance field defined on a deformable tetrahedral grid, where a mesh face will be extracted if two vertices of an edge in a tetrahedron have different signs of SDF values.
Here we transform the implicit SDF of canonical space to the deformable tetrahedral grid of deformed space via the SMPL-guided deformation process.
Given any vertexes $\mathbf{x_d}$ in the tetrahedral grid under the deformed space, we first find its corresponding point $\mathbf{x_c}$ in canonical space using Eq.~\ref{eq:warp} and query its signed distance value $d(\mathbf{x_d})$ from the canonical SDF as $d(\mathbf{x_d}) = d(\mathbf{x_c})$.
Then, we extract a triangular mesh of the generated human from the tetrahedral grid via the differentiable marching tetrahedra algorithm~\cite{shen2021deep}.

To achieve detailed geometric modeling of the generated humans, we further build a normal field on the extracted human mesh.
The normal field depicts the fine-grained geometry of the clothed humans by associating each surface point with a normal vector, whose direction and magnitude represent the orientation and curvature of the surface at each point. 
Following IGR~\cite{gropp2020implicit}, we first construct a canonical normal field by calculating the spatial gradient of the canonical signed distance fields as:
\begin{equation}
\label{eq:normal}
\begin{split}
 n(\mathbf{x_c}) =  \nabla_x d (\mathbf{x_c}),
\end{split}
\end{equation}
where $d(\mathbf{x_c})$ and $n(\mathbf{x_c})$ are the  signed distance value and  normal vector for  canonical point $\mathbf{x_c}$.
Similarly, we transform the normal field from the canonical space to the deformed space via the SMPL-guided deformation process.
For any points $\mathbf{x_d}$ at the extracted mesh surface in deformed space, we find $\mathbf{x_c}$ in canonical space via Eq.~\ref{eq:warp} and determine its surface normal vector $n(\mathbf{x_d})$ by:
\begin{equation}
\begin{aligned}
\label{eq:norm_transform}
n(\mathbf{x_d}) &= \left(\sum\limits_{i=1}^{N_j} s_i^* \cdot R_i (\theta, \beta) \right) \cdot n(\mathbf{x_c}), \\ 
\end{aligned}
\end{equation}
where $s_i^*$ and $R_i (\theta, \beta)$ are the skinning weights and rotation component of $B_i(\theta, \beta)$ in  Eq.~\ref{eq:warp}.
With the explict surface normal modeling, we can further render the extracted human mesh into a 2D normal map, enabling the model to learning realistic surface details from the normal maps of 3D scans.

\subsection{Efficient Differentiable Rendering}
\label{sec:rendering}
Existing 3D human GANs~\cite{noguchi2022unsupervised,bergman2022generative,zhang2022avatargen,hong2022eva3d} typically exploit implicit neural representation along with a neural volumetric rendering technique for 3D-aware human generation. However, such  neural volumetric rendering is computationally inefficient and GPU memory intensive, making them hardly  generate high-resolution images. Differently, our \nameofmethod{} adopts an explicit mesh representation for human modeling, which support highly efficient rasterizer-based rendering~\cite{laine2020modular}, and thus can generate images up to $1024^2$ resolution.

To render images, we first project the extracted mesh into a 2D mask and a coordinate map using the efficient rasterizer Nvdiffrast~\cite{laine2020modular} given the camera parameter $c$. Each pixel on the coordinate map stores the corresponding 3D coordinates on the mesh surface. Then, for every pixel on the coordinate map, we un-warp its corresponding 3D coordinate back to the canonical space and query the color value, yielding high-resolution RGB images (see Fig.~\ref{fig:framework}).

Additionally, our method can render normal maps directly from normal fields of the extracted meshes thanks to the differentiable  surface modeling and rendering techniques. This enables us to perform adversarial training on normal maps, which helps capture high-frequency geometric details such as wrinkles, hair, and faces from the 2D normal maps of 3D scans, as demonstrated in our experiments.

%-------------------------------------------------------------------------
\begin{figure*}[]
\centering
  \includegraphics[width=1\linewidth]{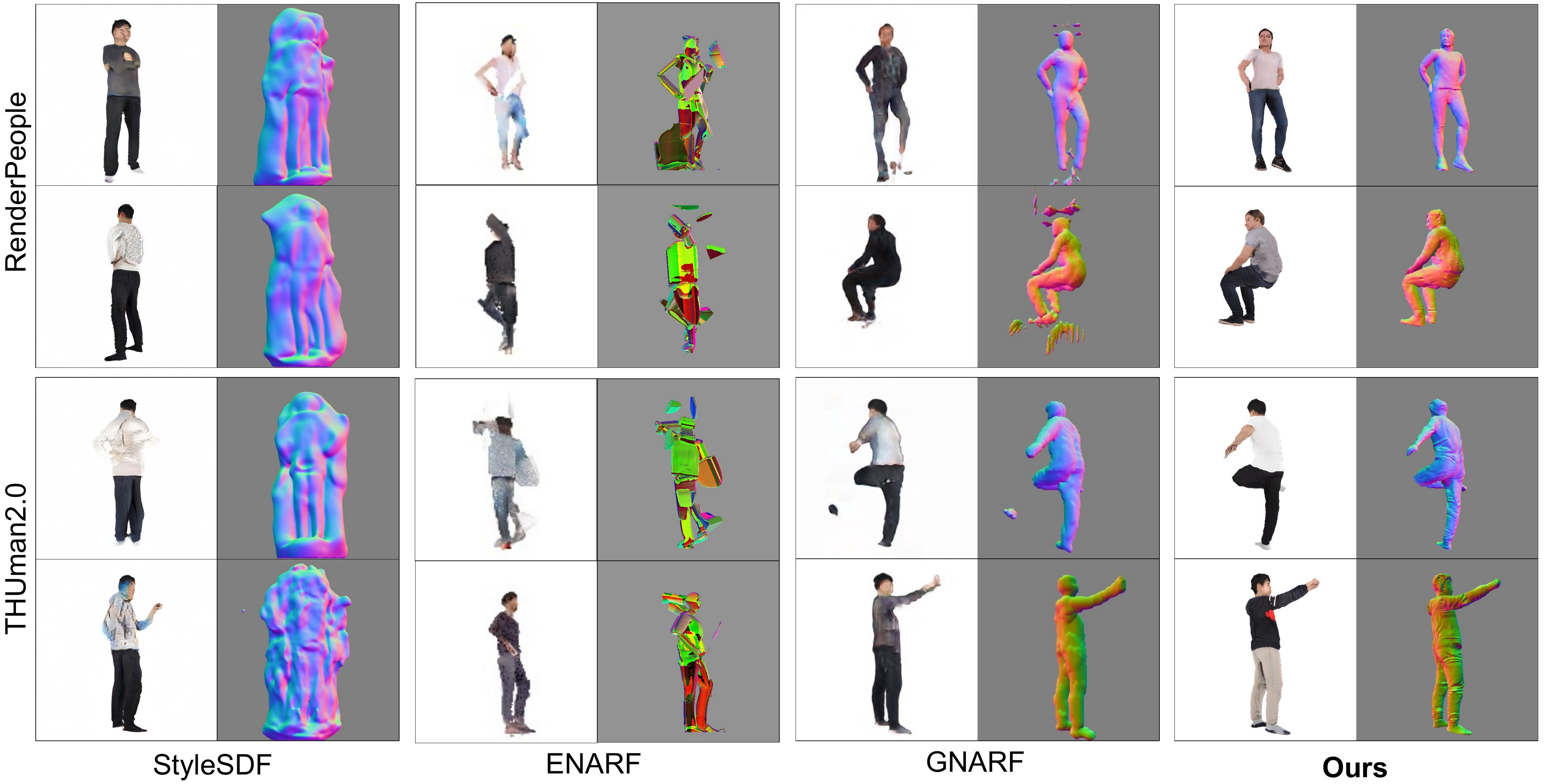}
  \vspace{-7mm}
  \caption{Quatitative comparison between StyleSDF~\cite{or2022stylesdf}, ENARF~\cite{noguchi2022unsupervised}, GNARF~\cite{bergman2022generative} and ours.
}
\vspace{-5mm}
\label{fig:sota}
\end{figure*}

\subsection{Adversarial Training}\label{sec:adv_training}
We train our \nameofmethod{} model from a collection of 2D human images with corresponding SMPL parameters $p=(\theta, \beta)$ and camera parameters $c$. 
We adopt a non-saturating GAN objective with $R1$ gradient penalty during the training process. 
To enable the model to learn 3D shapes, geometric details, and textures sufficiently, we use three StyleGAN2 discriminators~\cite{karras2020analyzing} to perform adversarial training on 2D masks, normal maps, and RGB images individually~\cite{gao2022get3d}. 
To encourage the surface normal to be an unit 2-norm, we apply Eikonal loss~\cite{gropp2020implicit,or2022stylesdf} on the surface of generated mesh:
\begin{equation}
\label{eq:eik_loss}
\begin{split}
 \mathcal{L}_{eik} = \sum\limits_{x_i}(|| \nabla_x d (x_{i})|| -1 )^2,
\end{split}
\end{equation}
where $x_i$ and $d(x_{i})$ denote the surface point and its signed distance value. 
To eliminate internal geometry and floating faces, we follow \cite{gao2022get3d,munkberg2022extracting} to regularize the signed distance values of DMTet~\cite{shen2021deep} using the cross-entropy loss:
\begin{equation}
\label{eq:ce_Loss}
\begin{split}
 \mathcal{L}_{ce} = \sum\limits_{i,j\in\mathbb{S}_e} H ( \sigma(d_i), sign(d_j ) ) + H ( \sigma (d_j ), sign(d_i) ),
\end{split}
\end{equation}
where $H$, $\sigma$,  $sign$ denote the binary cross-entropy, sigmoid,  sign function, respectively, and 
$\mathbb{S}_e$ is the set of edges where the signed distance values of two vertices have different signs.
Therefore, the overall loss is defined as:
\begin{equation}
\label{eq:total_Loss}
\begin{split}
 \mathcal{L}_{total} = \mathcal{L}_{adv} + \lambda_{eik} \mathcal{L}_{eik} + \lambda_{ce} \mathcal{L}_{ce},
\end{split}
\end{equation}
where $\lambda_{eik}=0.001$, $\lambda_{ce}=0.01$, and $\mathcal{L}_{adv}$ is the sum of adversarial losses on 2D masks, normal maps, and images.

\section{Experiments}
\noindent \textbf{Datasets.} 
We conduct experiments  on two high-quality 3D human scan datasets: 
THUman2.0~\cite{tao2021function4d} and RenderPeople~\cite{renderpeople}.
THUman2.0 contains 526 high-quality human scans with a diverse range of body poses captured by a dense DSLR rig, and provides official SMPL fitting results for each scan.
For RenderPeople, we utilize 1,190 scans with varying body shapes and textures to prepare the training images, and incorporated SMPL fits from AGORA dataset~\cite{patel2021agora}.
For every scan on these datasets, we employ Blender to render 100 RGB images, 2D silhouette masks, and normal maps with randomly-sampled camera poses.

\noindent  {\bf Evaluation Metrics.}
To evaluate the visual quality and diversity of the generated RGB images, we compute Frechet Inception Distance~\cite{heusel2017gans} between 50k generated RGB images and all real RGB images: FID$_{RGB}$. 
We evaluate the geometry quality of generated human avatars from 3 aspects: the quality of surface details, the correctness of generated poses, and the plausibility of generated depth.
First, to evaluate the quality of generated surface details, we measured Frechet Inception Distance~\cite{heusel2017gans} the normal maps:  FID$_{normal}$, between 50k generated normal maps and all real normal maps.
Second, to  measure the correctness of generated poses, we employ the Percentage of Correct Keypoints (PCK) metric, as used in previous animatable 3D human generation methods~\cite{noguchi2022unsupervised,bergman2022generative,hong2022eva3d,zhang2022avatargen}. 
To compute PCK, we first use a human pose estimation model~\cite{mmpose2020} to detect the human keypoints on both the generated and real images with the same camera and SMPL parameters. Then, we calculated the percentage of detected keypoints on the generated image within a distance threshold on the real image.
Additionally, we evaluate the depth plausibility by comparing the generated depths with the pseudo-ground-truth depth  estimated from the generated images by PIFuHD~\cite{saito2020pifuhd}.
To assess the rendering speed, we report  FPS running on a single NVIDIA V100 GPU.

\noindent  {\bf Baselines.}
We compare our method against both state-of-the-art 3D-aware image synthesis~\cite{or2022stylesdf,gao2022get3d,chan2022efficient} and 3D human generation~\cite{noguchi2022unsupervised,hong2022eva3d} methods. 
It is worth noting that 3D-aware image synthesis models cannot  control the human pose in the generated images. 
Since all compared baselines cannot produce body surface normal maps directly due to their lack of explicit surface modeling, 
we implement extra post-processing steps to obtain the normal maps from their generated meshes. 
Specifically,  we first reconstruct the 3D shapes from the density fields or SDFs using the  marching cube algorithm~\cite{lorensen1987marching}, and then render their normal maps using a PyTorch3D rasterizer~\cite{ravi2020pytorch3d}.

\subsection{Results}
\label{sec:mainresults}
\begin{table*}[]
\centering
\setlength{\tabcolsep}{0.8mm}
\begin{tabular}{l|c|c|c|cccc|cccc}
\Xhline{2\arrayrulewidth}
 \multirow{2}{*}{Method} & \multirow{2}{*}{Anim.} & \multirow{2}{*}{Res.} &  \multirow{2}{*}{FPS $\uparrow$} & \multicolumn{4}{c|}{THUman2.0} & \multicolumn{4}{c}{RenderPeople}  \\
&  & &  & FID$_{RGB}$ $\downarrow$ & FID$_{normal}$ $\downarrow$ & PCK $\uparrow$ & Depth $\downarrow$ & FID$_{RGB}$ $\downarrow$ & FID$_{normal}$ $\downarrow$ & PCK $\uparrow$ & Depth $\downarrow$\\
\toprule
\multirow{2}{*}{StyleSDF~\cite{or2022stylesdf}} & \multirow{2}{*}{\ding{55}} &$512^2$ & - & 80.45 & 266.97 & - & 1.14 & 68.65 & 259.22 & -  & 1.14 \\ 
 &  & $1024^2$ & & 94.72  &  273.43 & - & 1.41  & 81.27 &227.91 & -   & 1.33 \\
 \midrule
\multirow{2}{*}{GET3D~\cite{gao2022get3d}} & \multirow{2}{*}{\ding{55}} & $512^2$ & - & 73.71  & 136.53 & - & 1.16 & 37.95 & 124.28  & -  & 1.13 \\ 
 &  & $1024^2$ & - & 65.77 & 134.53 &-  & 0.92 & 42.63 & 106.84 & -  & 0.85 \\ 
 \midrule
\multirow{2}{*}{EG3D~\cite{chan2022efficient}} & \multirow{2}{*}{\ding{55}} & $512^2$ & - & 63.59  & 161.85 & - & 1.37 & 22.99 & 109.51 & -  & 1.12 \\ 
 &  & $1024^2$ &- & 75.70 &204.70  & - & 1.15 & 24.97 & 156.47  & -  & 1.04 \\  
 \midrule
ENARF~\cite{noguchi2022unsupervised} & $\checkmark$ & $128^2$ & 8 & 124.61  & 223.72 & 82.08 & 1.37
 & 108.59 & 205.26 &  75.66 & 1.26 \\ 
 \midrule
GNARF~\cite{bergman2022generative} & $\checkmark$ & $256^2$ & 8 & 68.31 & 166.62 & 94.28 & 1.44 & 55.07  & 132.35 & 93.28  & 1.62 \\ 
%\rowcolor{gray!25}
\midrule
\multirow{2}{*}{Ours} & \multirow{2}{*}{$\checkmark$} & $512^2$ & \textbf{17} &\textbf{13.54} & \textbf{22.31} & \textbf{99.61} & 0.83  & 12.65 & \textbf{34.58} & \textbf{99.12}  & 0.92 \\ 
 & & $1024^2$ & 14  & 17.91 & 55.02 & 99.39 & \textbf{0.82} & \textbf{11.77} & 58.57 & 98.99  & \textbf{0.73} \\ 
\Xhline{2\arrayrulewidth} 
\end{tabular}
\vspace{-2mm}
\caption{Quantitative comparisons with best results in \textbf{bold}. ``Anim." represents whether the method is animatable or not, and ``Res." denotes the image resolution.} 
\vspace{-3mm}
\label{table:comp}
\end{table*}

\noindent  {\bf Qualitative Results.}
We visualize the generated RGB images and normal maps for qualitative comparison, as shown in  Fig.~\ref{fig:sota}.
From the results, we can make the following observations.  StyleSDF~\cite{or2022stylesdf} produces distorted body shapes due to its lack of an explicit human body representation.
Even though ENARF~\cite{noguchi2022unsupervised} models the pose prior as a skeletal distribution, it still struggles to generate human avatars with correct target poses due to the inaccurate body deformation modeling. 
GNARF~\cite{bergman2022generative} can render reasonable RGB images but suffers from noisy geometries. 
Besides, it generates ``floating noises" outside the human body with large pose articulations. 
Additionally, the geometries of generated human bodies are coarse and over-smoothed, lacking geometric details like clothes and hairs. 
Due to the high rendering cost and representation limitation, the rendered images from  ENARF~\cite{noguchi2022unsupervised} and GNARF~\cite{bergman2022generative} have relatively low resolutions, resulting in low-quality rendering results.
In contrast, as shown in the right column of Fig.~\ref{fig:sota}, our method generates significantly better human avatars with detailed body geometries, even under large pose articulations like ``sitting" and ``single-leg standing".
Please refer to the Appendix for more visualization results (\eg, animation and multi-view videos).

%-------------------------------------------------------------------------

\noindent  {\bf Quantitative Evaluations.}
As shown in Tab.~\ref{table:comp}, \nameofmethod{} consistently outperforms all the baseline methods~\cite{or2022stylesdf,gao2022get3d,chan2022efficient,noguchi2022unsupervised,bergman2022generative} on both datasets~\cite{renderpeople,tao2021function4d} \wrt all  the metrics.
We observe that 3D-aware image generation models (StyleSDF~\cite{or2022stylesdf}, GET3D~\cite{gao2022get3d}, and EG3D~\cite{chan2022efficient})  struggle to achieve reasonable FID scores on the THUman2.0~\cite{tao2021function4d} due to its complexity and diverse poses.
ENARF~\cite{noguchi2022unsupervised} and GNARF~\cite{bergman2022generative} can only generate relatively low-resolution images,\ie, resolution of $128^2$ and $256^2$.
In contrast, our model produces high-resolution human images ($512^2$ and $1024^2$) with superior visual quality (FID$_{RGB}$), geometry quality (FID$_{normal}$), pose controllability (PCK), and depth plausibility (Depth). 
In terms of inference speed, \nameofmethod{} generates $512^2$ images at 17 FPS and $1024^2$ images at 14 FPS, while ENARF~\cite{noguchi2022unsupervised} and GNARF~\cite{bergman2022generative} operate at 8 FPS for $128^2$ and $256^2$ images, respectively, verifying the efficiency of our method.

%-------------------------------------------------------------------------
\begin{figure}[]
\centering
\vspace{-2mm}
  \includegraphics[width=1\linewidth]{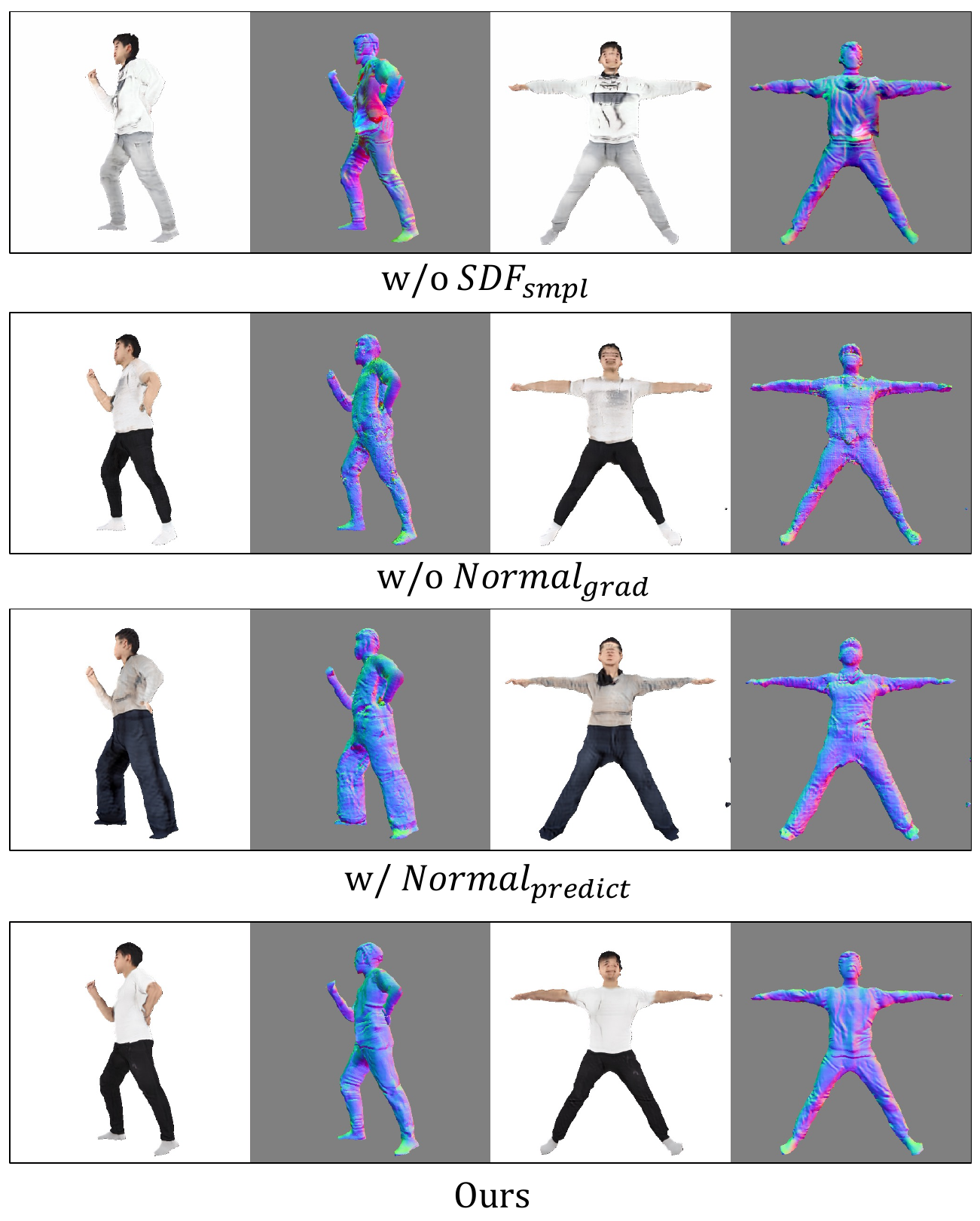}
  \vspace{-7mm}
  \caption{\textbf{Abalation on the model designs.}
  We visualize the human avatars with both deformed and canonical poses.
}
\label{fig:ablation}
\vspace{-3mm}
\end{figure}

%-------------------------------------------------------------------------

\subsection{Ablation Studies}
\label{sec:ablation}
\noindent  {\bf Geometry Modeling Scheme.}
% SDF: raw, res
% normal supervision: wo / w
We further conduct experiments on THUman2.0~\cite{tao2021function4d} dataset to analyze the impact of different model designs. 
To begin with, we investigate the effects of SDF modeling, and find that directly predicting the sign distance value without incorporating the geometry prior of the SMPL template led to noticeable artifacts, \eg, strange geometries in the waist region, as illustrated in the first row of Fig.~\ref{fig:ablation}. 
We then examine the effect of the normal field.
The second row of Fig.~\ref{fig:ablation} demonstrates that the generated human bodies contain noises and holes when the gradient-based normal field modeling is not included. 
We also explore an alternative method of modeling the surface normal by predicting the normals from an MLP~\cite{tiwari2021neural,lin2022magic3d}. 
However, as shown in the third and fourth row of Fig.~\ref{fig:ablation}, we observe that the geometry quality of the predicted normal modeling method is inferior to our gradient-based approach. 
To further assess the effects of our geometry modeling scheme, we conduct quantitative experiments, the results of which are presented in Tab.~\ref{table:ablation}.
Both the quantitative and qualitative findings demonstrate that the normal field modeling and normal map supervision contribute to better geometry and appearance quality.

\begin{table}[]
\centering
% \small
\setlength{\tabcolsep}{0.4mm}
\begin{tabular}{l|cccc}
\Xhline{2\arrayrulewidth}
Method &   FID$_{RGB}$ $\downarrow$ & FID$_{normal}$ $\downarrow$ & PCK $\uparrow$ & Depth $\downarrow$ \\
\toprule
w/o SDF$_{smpl}$ & 15.80 &  29.52  & 99.47 & 0.92 \\ 
w/o Normal$_{grad}$ &  20.23   & 63.49 & 99.25 &   0.96  \\ 
w/ Normal$_{predict}$ &  16.93  & 89.38 & 99.11 &  0.88 \\ 
Ours & \textbf{13.54}  & \textbf{22.31} & \textbf{99.61} &   \textbf{0.83} \\ 
\Xhline{2\arrayrulewidth} 
\end{tabular}
\caption{Ablation studies on geometry modeling scheme.} 
\label{table:ablation}
\vspace{-1mm}
\end{table}

\subsection{Applications}
\label{sec:applicaion}

\begin{figure}[]
\centering
  \includegraphics[width=1\linewidth]{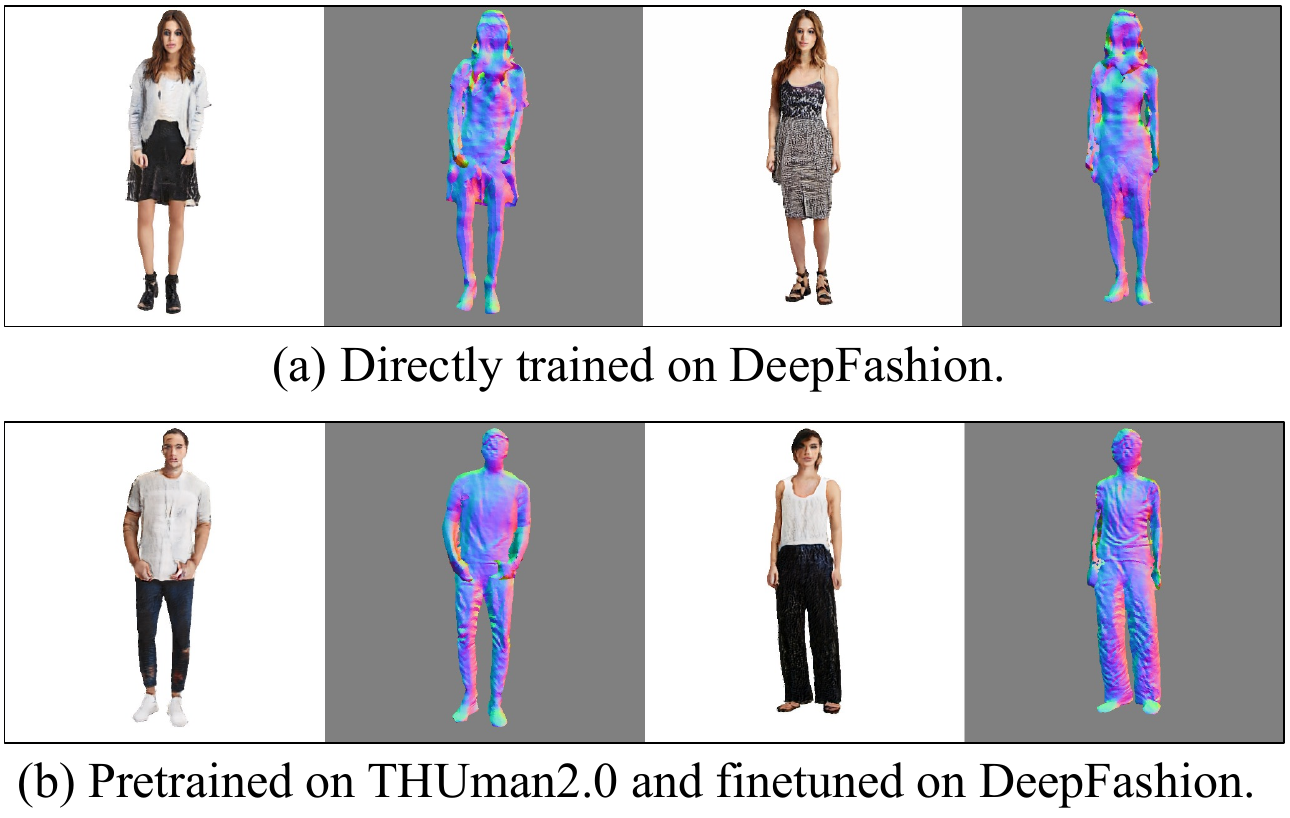}
  \vspace{-4mm}
  \caption{\textbf{Transfer learning.}
  Compared to directly training on DeepFashion dataset~\cite{liu2016deepfashion}, we find that pretraining on THUman2.0~\cite{tao2021function4d} leads to much better geometries.
}
\label{fig:transfer}
\vspace{-4mm}
\end{figure}

\noindent  {\bf Transfer Learning.}
Benefiting from the explicit mesh representation, our method is ready for transfer learning.
In practice, normal maps may not be available for  in-the-wild  datasets, such as DeepFashion~\cite{liu2016deepfashion}.
We observe that training directly on these datasets leads to degenerate results due to insufficient geometric supervision by learning from RGB images alone (see the first row of Fig.~\ref{fig:transfer}). 
To improve the generated geometries, one possible solution is to inherit the human geometry knowledge by learning from the 3D human datasets~\cite{liu2016deepfashion} via transfer training.
We first pretrain the model on the THUman2.0~\cite{tao2021function4d} containing 2D normal maps, and then fine-tune on the DeepFashion~\cite{liu2016deepfashion}, leveraging both the rich geometry information of the 3D human dataset and the diverse texture information of the 2D fashion dataset.
From Fig.~\ref{fig:transfer}, we observe that the transfer learning significantly improves the geometries on 2D fashion dataset~\cite{liu2016deepfashion}.

%-------------------------------------------------------------------------

\begin{figure}[]
\centering
% \vspace{-2mm}
  \includegraphics[width=1\linewidth]{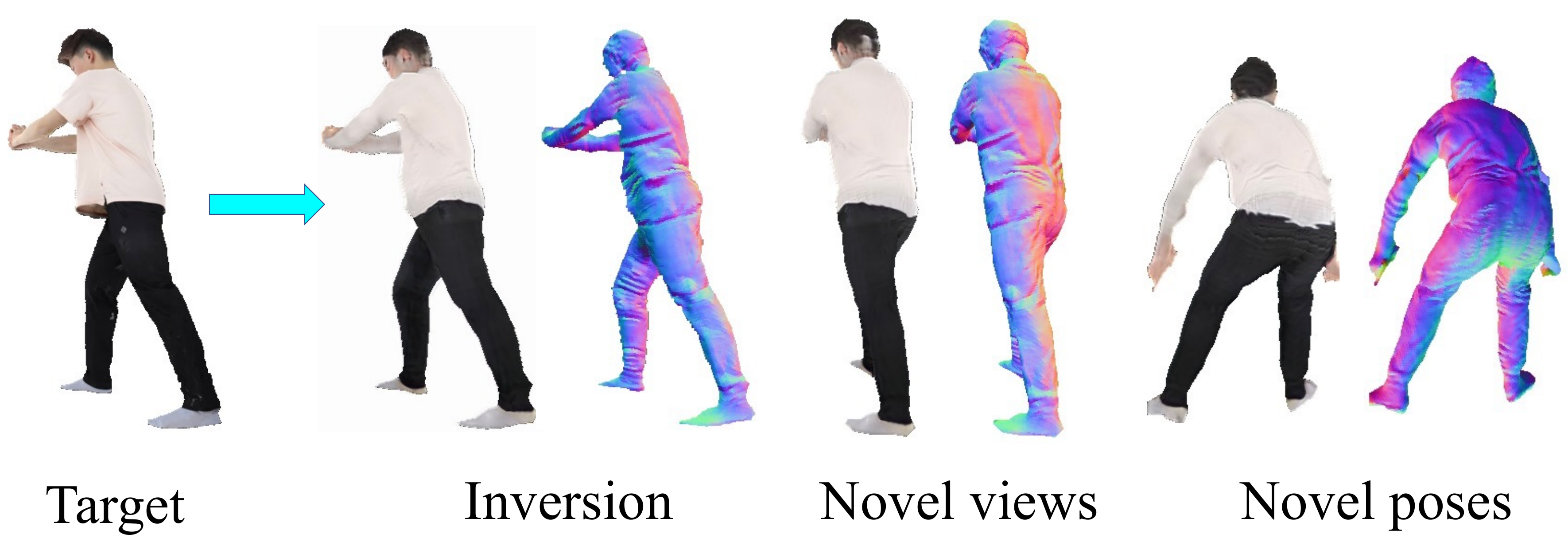}
  \caption{\textbf{Inversion.}  Given a target image, we reconstruct
    its 3D human avatar and manipulate to novel camera views and novel poses. 
}
\label{fig:inverse}
\vspace{-2mm}
\end{figure}

\noindent  {\bf Single-view 3D Reconstruction and Manipulation.}
 \nameofmethod{} enables the creation of full-body human avatars using a single-view portrait image.
We adopt the optimization approach proposed in~\cite{karras2019style} to fit the given image. 
The optimization process involves using a frozen model with camera pose and SMPL parameters  estimated from the portrait image, and minimizing the mean squared error (MSE) and perceptual loss~\cite{zhang2018unreasonable} between the target and generated images.
Once the optimization process is complete, the reconstructed portrait can be manipulated to different camera views and human poses, based on certain controlling signals as shown in Fig.~\ref{fig:inverse}.

%-------------------------------------------------------------------------
%-------------------------------------------------------------------------

%-------------------------------------------------------------------------
\begin{figure}[]
\centering
  \includegraphics[width=0.9 \linewidth]{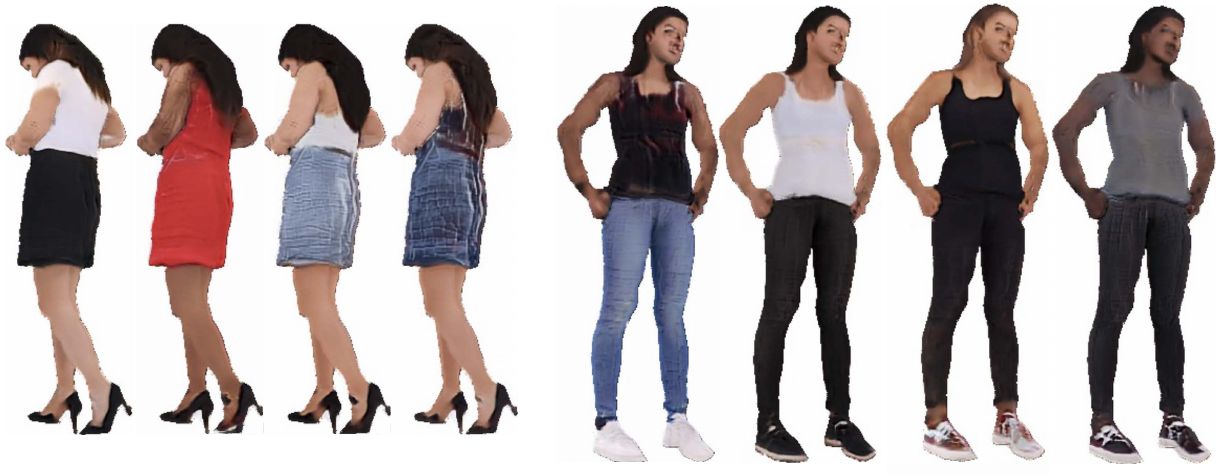}
  \caption{\textbf{Retexturing.}
  GETAvatar supports changing the textures of generated humans while maintaining the underlying geometries of bodies.
}
\label{fig:retexture}
\vspace{-2mm}
\end{figure}
%-------------------------------------------------------------------------

\noindent  {\bf Re-texturing.}
 \nameofmethod{} utilizes two separate triplane branches to represent the geometry and texture,  allowing for the disentanglement of geometry and texture.
This disentanglement enables the application of re-texturing by combining a shared geometry latent with various texture codes.
As shown in Fig.~\ref{fig:retexture}, our approach can alter the textures of the generated humans by modifying different texture codes while preserving the underlying body geometries.

\section{Conclusion}
In this work, we introduce  \nameofmethod{}, a 3D-aware generative model that directly generates explicit textured 3D meshes for controllable full-body human avatars.
To enrich realistic surface details from 2D normal maps, 
we perform differentiable surface modeling by extracting the underlying surface as a 3D mesh and further building a normal field to depict the surface details.
The explicit mesh representation enables us to achieve high-resolution rendering efficiently when combined with a rasterization-based renderer.
Extensive experiments that GETAvatar achieves the state-of-the-art performance on 3D-aware human generation in terms of visual quality, geometry quality, and inference speed.

\noindent  {\bf Limitations.} 
Although our method can generate high-fidelity animatable 3D human avatars, there is still room for improvement.
One limitation is that it lacks the ability to control fine motions of the human avatars, such as changes in facial expressions.
 To address this issue, we could consider using more expressive 3D human models, such as SMPL-X~\cite{SMPL-X:2019}, as the guidance of deformation.
For ethical considerations, the proposed method could be misused for generating fake imagery of real people, and we do not condone using our model with the intent of spreading disinformation.

{\small
\bibliographystyle{ieee_fullname}
\bibliography{egbib}
}

%-------------------------------------------------------------------------
\clearpage

\section*{\Large\textbf{Appendix}}
\textit{
In the appendix, we first present the implementation details in Sec.~\ref{sec:implementation}.
Second, we provide the experimental details in Sec.~\ref{sec:experimental}.
Finally, we show more visualization results in Sec.~\ref{sec:additional}. 
% \textcolor{red}{add project page link}
Please also refer to the project page for video results.
Our project page: \url{https://getavatar.github.io/}.}

\let\thefootnote\relax\footnotetext{*Equal contribution.}
%%%%%%%%%%%%%%%%%%%%%%%%%%%%%%%%%%%%%%%%%%%%%%%%%%%%%%%%%%%%%%%%%%%%%%%%%%%%%%%%%%%%%%%%%%%%%%%%%%%
\section{Implementation Details}
\label{sec:implementation}

\subsection{Network Architectures}
\label{sec:arch}
\noindent \textbf{Triplane.}
Following EG3D~\cite{chan2022efficient} and GET3D~\cite{gao2022get3d}, we adopt the StyleGAN2~\cite{karras2020analyzing} generator to generate the triplane representation.
Specifically, the backbone (StyleGAN2~\cite{karras2020analyzing} generator) produces two triplanes: the texture triplane and the geometry triplane.
We employ two conditional layers for each style block to generate geometry features and texture features separately~\cite{li2021semantic,gao2022get3d}.
For each triplane, the backbone outputs a 96-channel output feature map that is then reshaped into three axis-aligned feature planes, each of shape $256 \times 256 \times 32$.

\noindent \textbf{Mapping Network.}
Both the geometry and texture mapping network are  8-layer MLPs network with leakyReLU as the activation function, and the dimension of the hidden layers is 512. We sample the input latent code $z_{geo} \in \mathbb{R}^{512}$ and $z_{tex}  \in \mathbb{R}^{512}$ from a 512-dimensional standard Gaussian distribution. 

\noindent \textbf{Discriminator.}
We use 3 StyleGAN2-based~\cite{karras2020analyzing} discriminators to perform adversarial training on RGB images, 2D masks, and normal maps, respectively.
Following EG3D~\cite{chan2022efficient}, we condition all the discriminators on the camera parameters by modulating the blocks of the discriminator via a mapping network.

\noindent \textbf{SMPL-guided Deformation.}
SMPL defines a deformable mesh $\mathcal{M}(\beta, \theta) = (\mathcal{V}, \mathcal{S})$, where $\theta$ denotes the pose parameter,  $\beta$ represents the shape parameter, $\mathcal{V}$ is the set of  $N_{v} = 6890$ vertices, and $\mathcal{S}$ is the set of linear blend skinning weights assigned for each vertex.
The template mesh of SMPL can be deformed by linear blend skinning~\cite{lewis2000pose} with $\theta$  and $\beta$.
Specifically, the linear blend skinning process can transform a vertex from the canonical pose to the target pose by the weighted sum of skinning weights that represent the influence of each bone and transformation matrices.
In this work, we generalize the linear blend skinning process~\cite{lewis2000pose} of the SMPL model from the coarse naked body to our generated clothed human. 
The core idea is to associate each point with its closest vertex on the deformed SMPL mesh  $\mathcal{M}( \theta,\beta)$,  assuming they undergo the same kinematic changes between the deformed and canonical spaces. Specifically, for a point $\mathbf{x_d}$ in the deformed space, we first find its nearest vertex $v^*$ in the SMPL mesh.
Then we use the skinning weights of $v^*$ to un-warp $\mathbf{x_d}$ to  $\mathbf{x_c}$ in the canonical space:
\begin{equation}
\begin{aligned}
\label{eq:warp}
\mathbf{x_c} &= \left(\sum\limits_{i=1}^{N_j} s_i^* \cdot B_i (\theta, \beta) \right)^{-1} \cdot \mathbf{x_d}, 
\end{aligned}
\end{equation}
where $N_j = 24 $ is the number of joints, $s_i^*$ is the skinning weight of vertex $v^*$  \wrt the $i$-th joint, 
$B_i (\theta, \beta)$ is the bone transformation matrix of join $i$.
Therefore, for any point $\mathbf{x_d}$ in the deformed space, we can determine the SDF value $d(\mathbf{x_d})$, color $c(\mathbf{x_d})$, and normal $n(\mathbf{x_d})$ as:
\begin{equation}
\begin{aligned}
\label{eq:transform}
d(\mathbf{x_d}) &= d(\mathbf{x_c}), \ \ c(\mathbf{x_d}) = c(\mathbf{x_c}), \\
n(\mathbf{x_d}) &= \left(\sum\limits_{i=1}^{N_j} s_i^* \cdot R_i (\theta, \beta) \right) \cdot n(\mathbf{x_c}), \\ 
\end{aligned}
\end{equation}
where $d(\mathbf{x_c})$, $c(\mathbf{x_c})$, $n(\mathbf{x_d})$ are the SDF value, color, and normal at the point $\mathbf{x_c}$ in the canonical space and $R_i (\theta, \beta)$ is the rotation component of $B_i(\theta, \beta)$.

\noindent \textbf{Differentiable Marching Tetrahedra.}
To explicitly model the body surface, we extract a triangular mesh of the generated human from the tetrahedral grid via the differentiable marching tetrahedra algorithm~\cite{shen2021deep}.
For the tetrahedra grid, the marching tetrahedra algorithm~\cite{shen2021deep} finds the surface boundary based on the sign of the signed distance value for vertices within each tetrahedron.
If two vertices $i$ and $j$ in the edge of a tetrahedron have opposite signs for the signed distance value ($sign(d_i) \neq sign(d_j) $), we can determine the mesh face vertice by a linear interpolation between vertices $i$ and $j$.

% \textcolor{red}{Add humannerf related introduction?}

\subsection{Training Protocol}
\label{sec:pro}

\noindent \textbf{Hyperparameters.}
We use Adam optimizer~\cite{kingma2014adam} with $\beta_1 = 0$, $\beta_2 = 0.99$, and the batch size of 32 for optimization.
The learning rate is set to $0.002$ for both the generator and the discriminator.
Following StyleGAN2~\cite{karras2020analyzing}, we use lazy regularization to stabilize the training process by applying R1 regularization to discriminators every 16 training steps.
Here we set the regularization weight to 10 for THUman2.0~\cite{tao2021function4d} and 20 for RenderPeople~\cite{renderpeople}.
For the loss function, we set $\lambda_{eik}=0.001$ for the eikonal loss, and $\lambda_{ce}=0.01$ for the cross-entropy loss of SDF regularization.

\noindent \textbf{Runtime Analysis.}
At training time, for images at $512^2$ resolution, we train the model on 8 NVIDIA Tesla V100 GPUs using a batch size of 32 for 1 day.
For images at $1024^2$ resolution, the models are trained on 8 NVIDIA A100 GPUs for 1 day, with a batch size of 32.
At test time, we evaluate the rendering speed in frames per second (FPS) at different resolutions.
In particular,  our model runs at $512^2$ resolution with 17FPS and  $1024^2$ resolution with 14FPS on a single NVIDIA Tesla V100 GPU.

\section{Experimental Details}
\label{sec:experimental}

\subsection{Datasets}
We conduct experiments on two high-quality 3D human scan datasets: THUman2.0~\cite{tao2021function4d} and RenderPeople~\cite{renderpeople}.
For every scan on these datasets, we render 100 RGB images, 2D silhouette masks, and normal maps with randomly-sampled camera poses.
Specifically, we sample the pitch and yaw of the camera pose from a uniform distribution with the horizontal standard deviation of 2$\pi$ radians and the vertical standard deviation of 0.1 radians.
Besides, we use a fixed radius of 2.3 and the fov angle of 49.13$^\circ$  for all camera poses.
For the SMPL parameters, we adopt the official provided SMPL fitting results\footnote{\url{https://github.com/ytrock/THuman2.0-Dataset}}  for THUman2.0~\cite{tao2021function4d}, and the SMPL fitting results\footnote{\url{https://agora.is.tue.mpg.de/}}  provided by  AGORA dataset~\cite{patel2021agora} for RenderPeople~\cite{renderpeople}.

\subsection{Evaluation Metrics}
\subsubsection{Texture Evaluation}
To evaluate the visual quality and diversity of the generated RGB images, we compute Frechet Inception Distance~\cite{heusel2017gans} between 50k generated RGB images and all real RGB images: FID$_{RGB}$. 
We adopt the FID implementation provided in the StyleGAN3 codebase\footnote{\url{https://github.com/NVlabs/stylegan3}}.

\subsubsection{Geometry Evaluation}
We evaluate the geometry quality of generated human avatars from 3 aspects: the quality of surface details, the correctness of generated poses, and the plausibility of generated depth.
First, to evaluate the quality of generated surface details, we measured Frechet Inception Distance~\cite{heusel2017gans} the normal maps:  FID$_{normal}$, between 50k generated normal maps and all real normal maps.
We adopt the widely-used implementation\footnote{\url{https://github.com/NVlabs/stylegan3}} of FID with a pretrained Inception v3 feature extractor\footnote{\url{https://api.ngc.nvidia.com/v2/models/nvidia/research/stylegan3/versions/1/files/metrics/inception-2015-12-05.pkl}}.
Second, to  measure the correctness of generated poses, we employ the Percentage of Correct Keypoints (PCK) metric, as used in previous animatable 3D human generation methods\footnote{\url{https://github.com/nogu-atsu/ENARF-GAN}}
\cite{noguchi2022unsupervised,bergman2022generative,hong2022eva3d,zhang2022avatargen}. 
To compute PCK, we first use a human pose estimation model\footnote{\url{https://github.com/open-mmlab/mmpose}} to detect the human keypoints on both the generated and real images with the same camera and SMPL parameters. Then, we calculated the percentage of detected keypoints on the generated image within a distance threshold on the real image.
Additionally, we evaluate the depth plausibility by comparing the generated depths with the pseudo-ground-truth depth  estimated from the generated images by PIFuHD\footnote{\url{https://github.com/facebookresearch/pifuhd}}~\cite{saito2020pifuhd}.

\subsubsection{Baselines}
\begin{figure}[h]
\centering
\vspace{-3mm}
\includegraphics[width=0.8\linewidth]{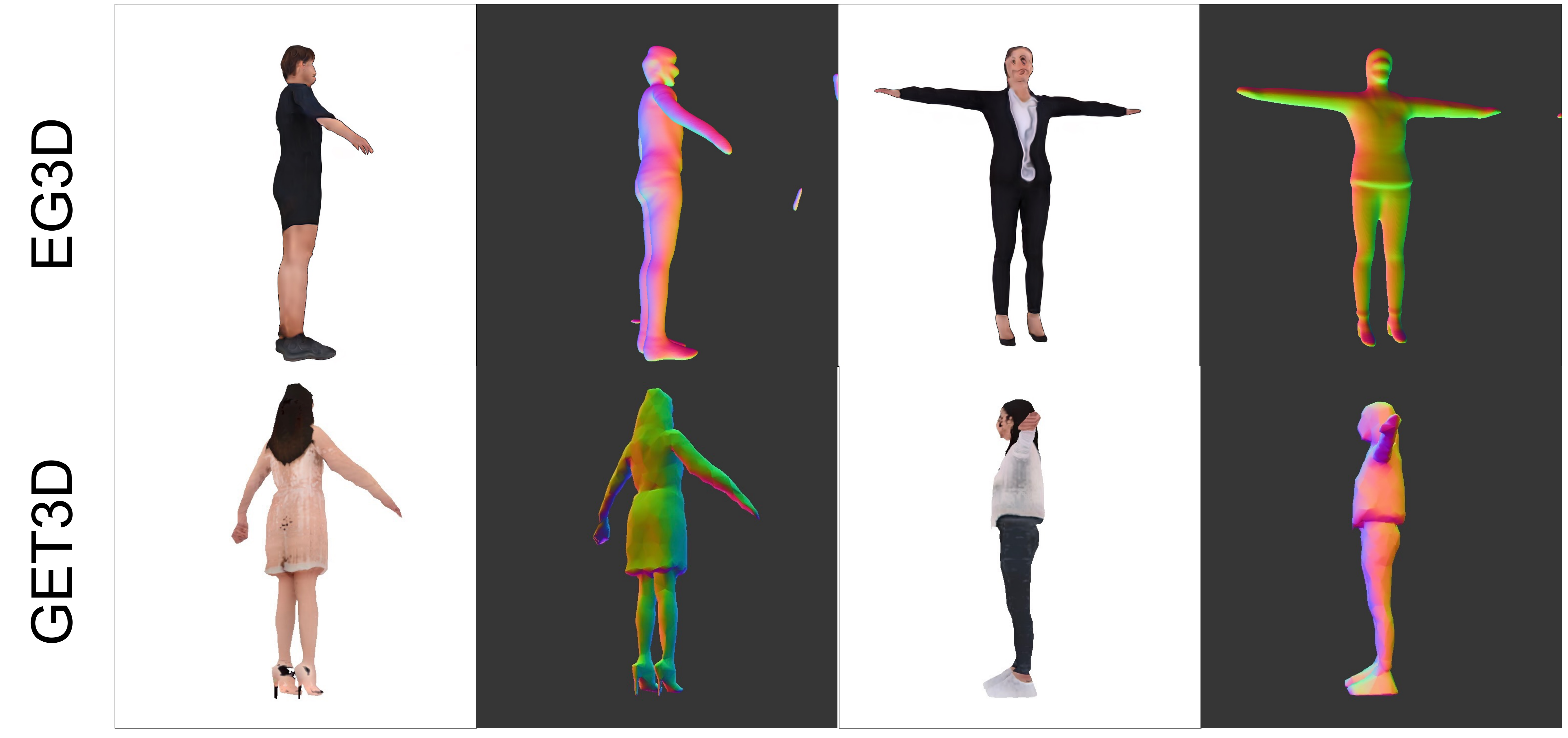}
\vspace{-4mm}
\label{fig:eg3d}
\end{figure}

When training the 3D-aware image synthesis~\cite{or2022stylesdf,gao2022get3d,chan2022efficient} models, we follow the official implementations to train the model with only parameters.
We also visualize the generated RGB images and normal maps in Fig.~\ref{fig:eg3d}.

\section{Additional Results} \label{sec:additional}
\begin{figure*}
{
      \includegraphics[width=1.0\linewidth]{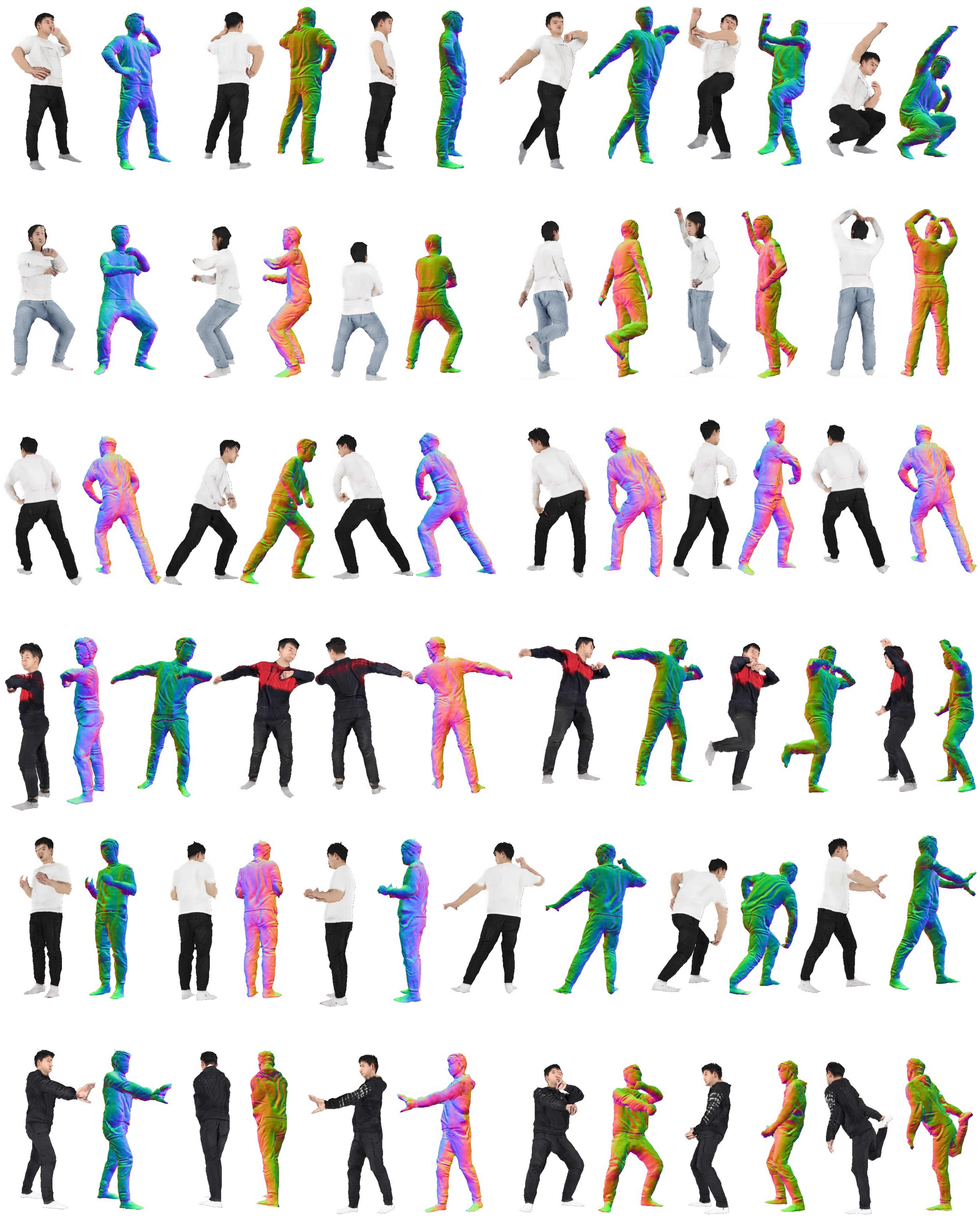}
}
 \vspace{-5mm}
\caption{Images synthesized by \nameofmethod{} on the THUman2.0~\cite{tao2021function4d} dataset.}
\label{fig:thu_result}
\end{figure*}

\begin{figure*}
{
      \includegraphics[width=0.5\linewidth]{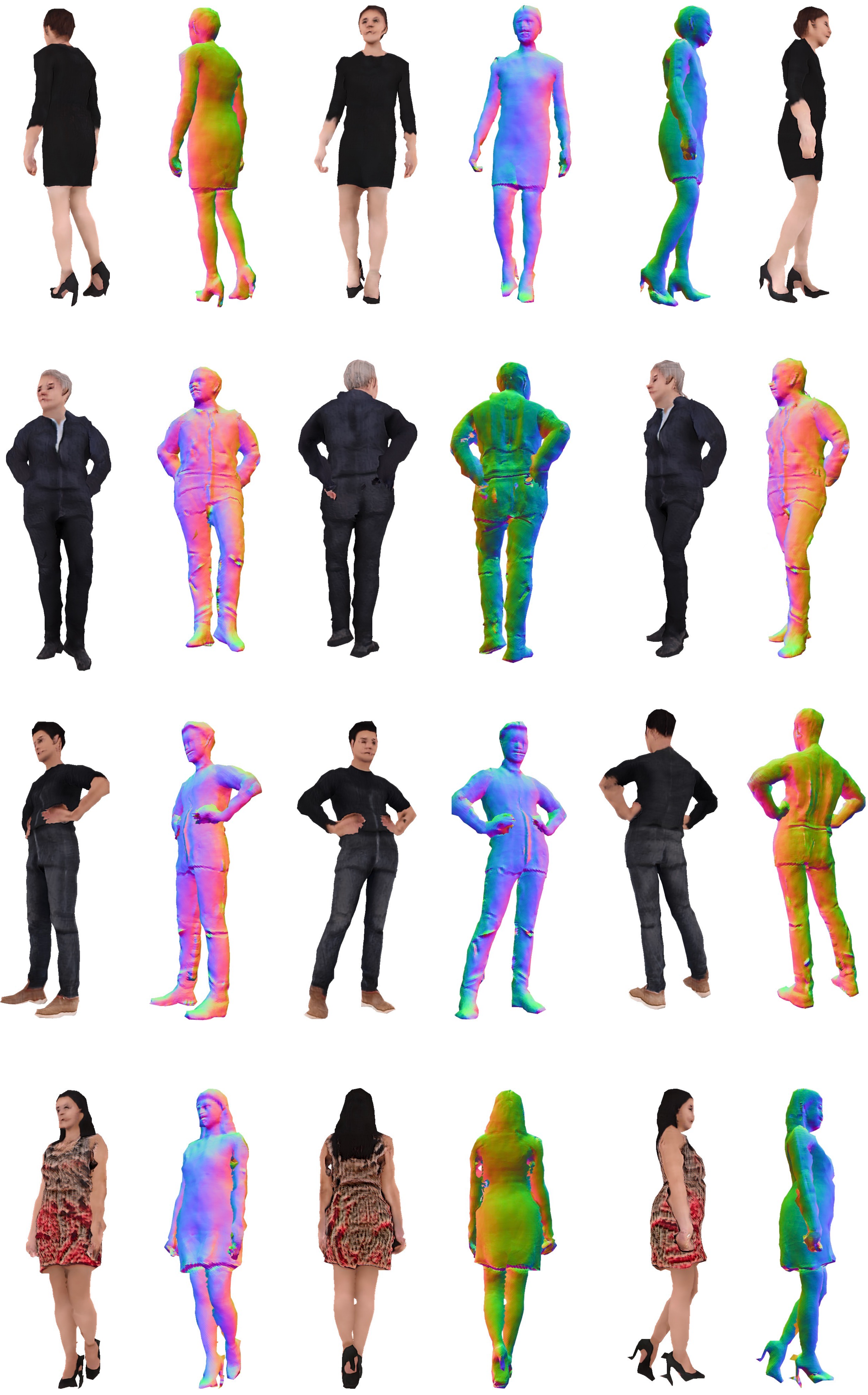}
       \includegraphics[width=0.5\linewidth]{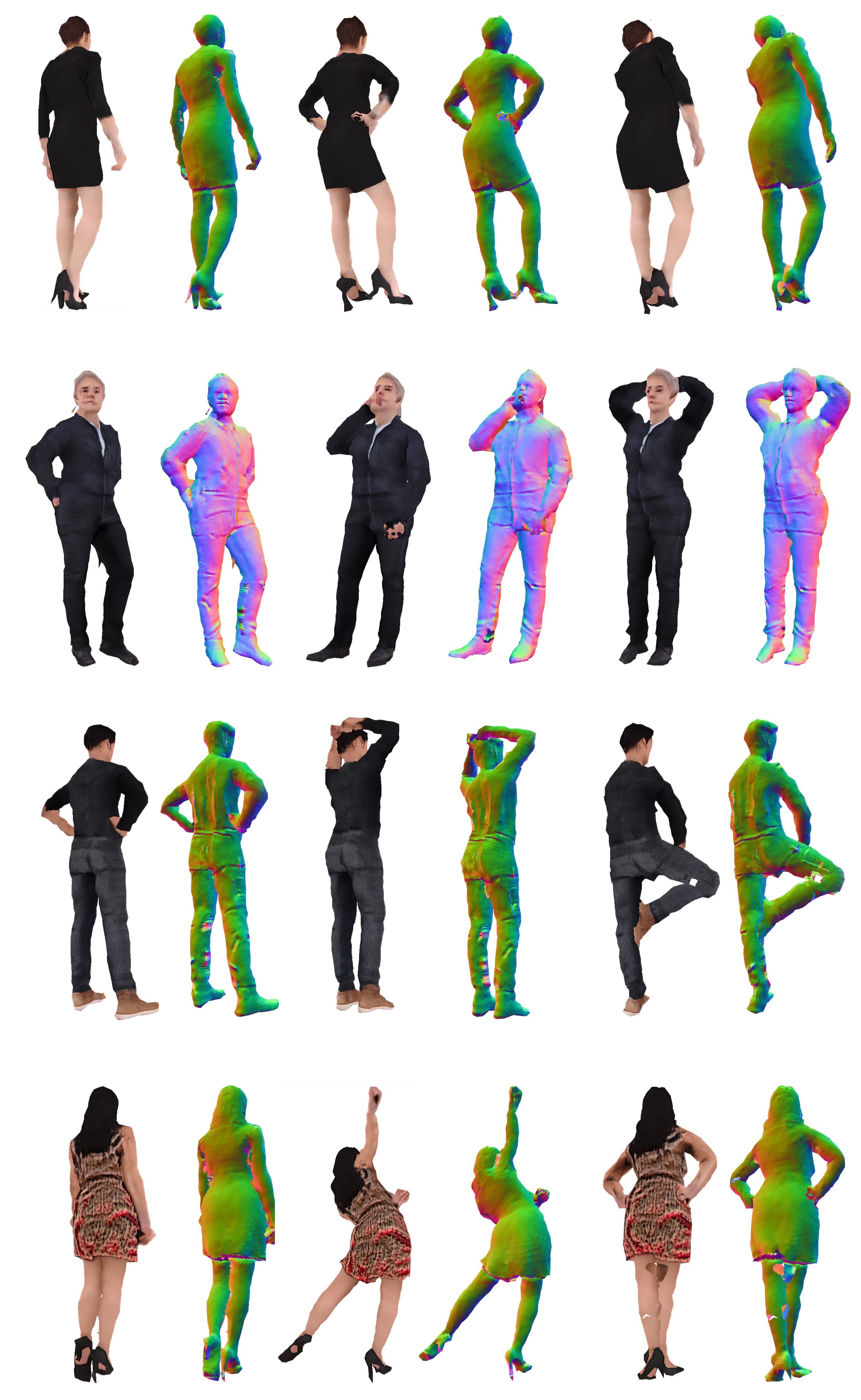}
       \includegraphics[width=0.5\linewidth]{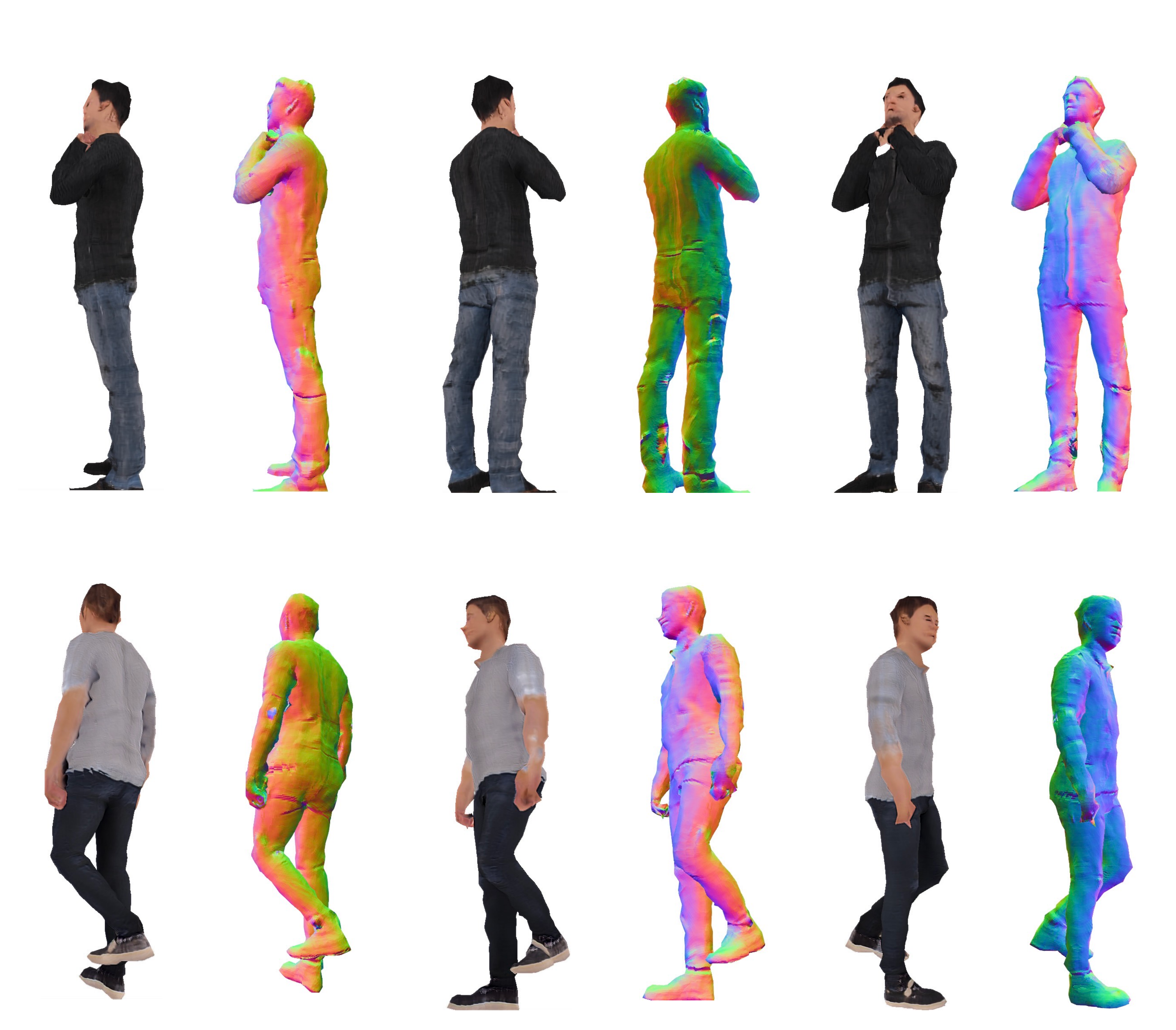}
       \includegraphics[width=0.5 \linewidth]{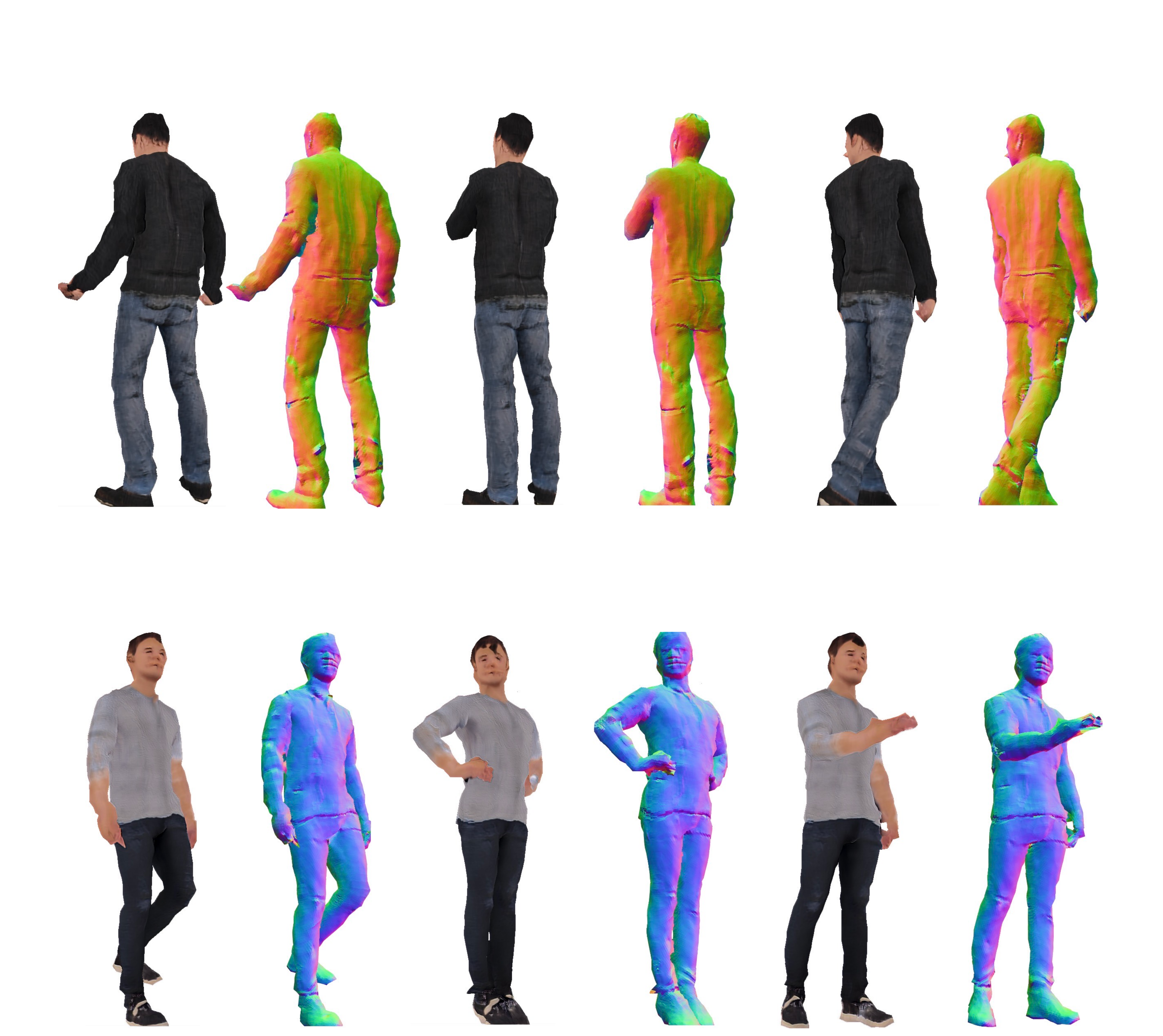}
}
 \vspace{-5mm}
\caption{Images synthesized by \nameofmethod{} on RenderPeople~\cite{renderpeople}.}
\label{fig:rp_result}
\end{figure*}

\begin{figure*}
{
      \includegraphics[width=1.0\linewidth]{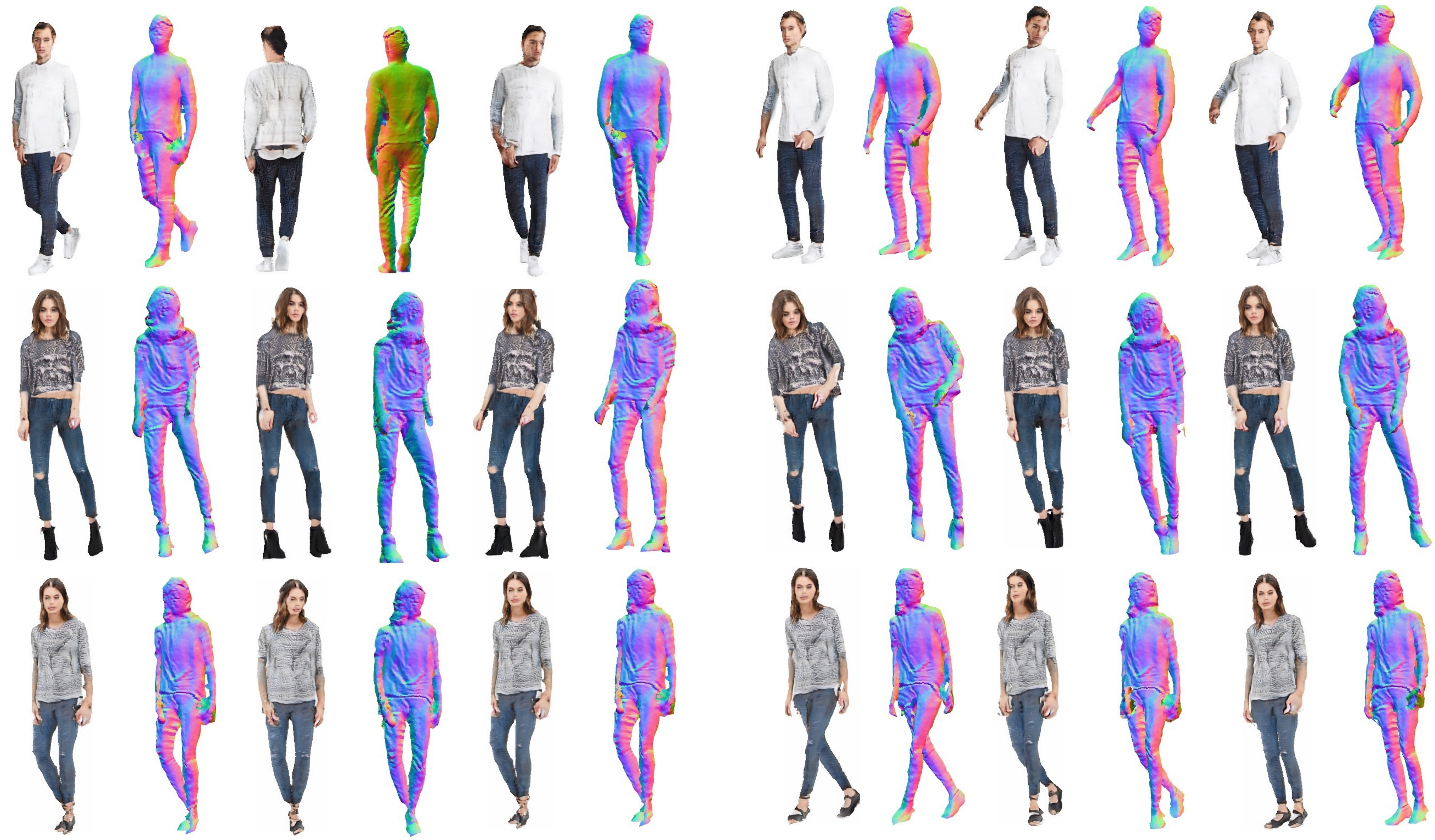}
}
 \vspace{-5mm}
  \caption{Images synthesized by \nameofmethod{} on  DeepFashion~\cite{liu2016deepfashion}. }
  \vspace{-1mm}
\label{fig:df_result}
\end{figure*}

\begin{figure*}[t]
{
      \includegraphics[width=1.0\linewidth]{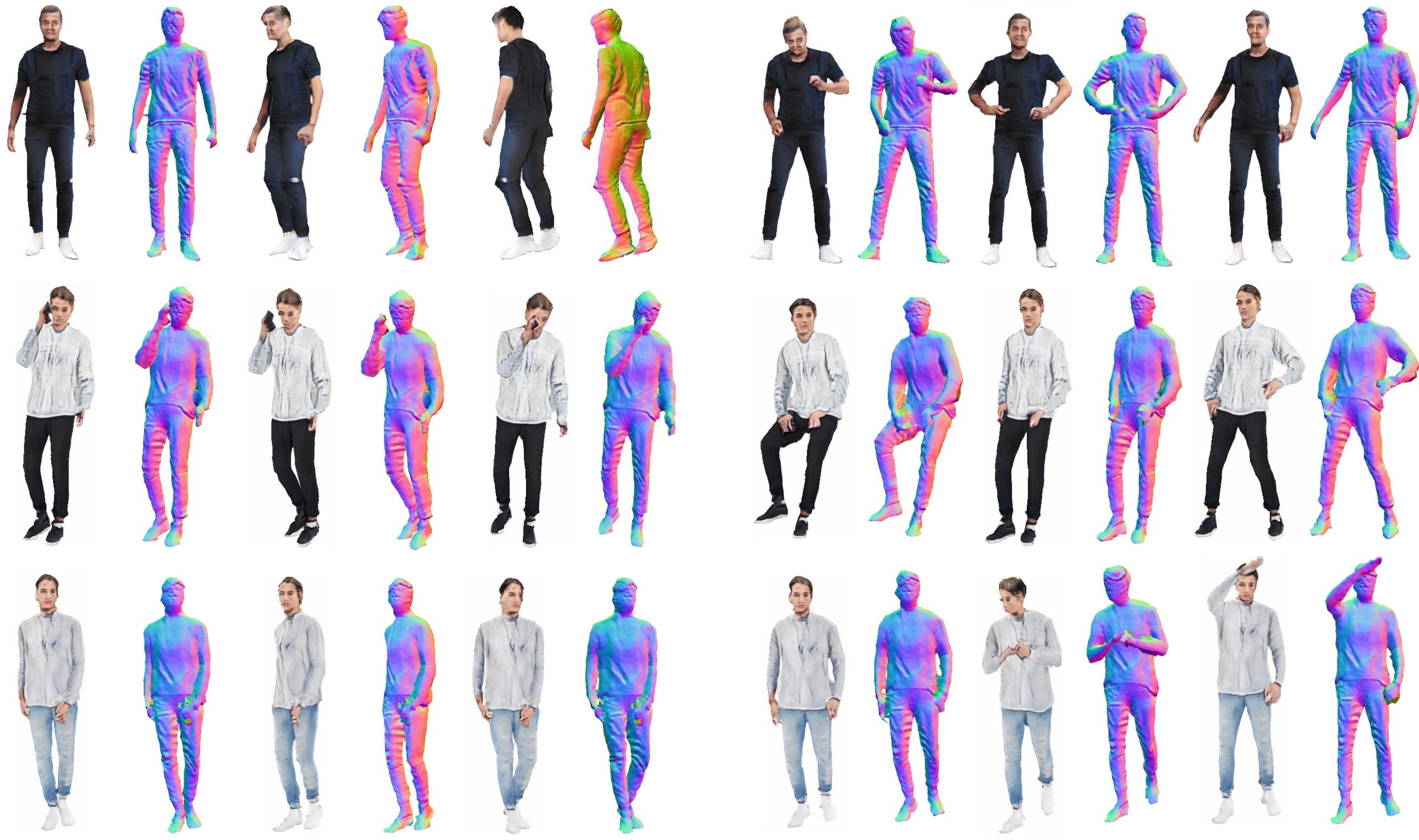}
}
\vspace{-5mm}
  \caption{Images synthesized by \nameofmethod{} on SHHQ~\cite{fu2022stylegan}. }
\label{fig:shhq_result}
\end{figure*}

\begin{figure*}[ht]
{
      \includegraphics[width=1.0\linewidth]{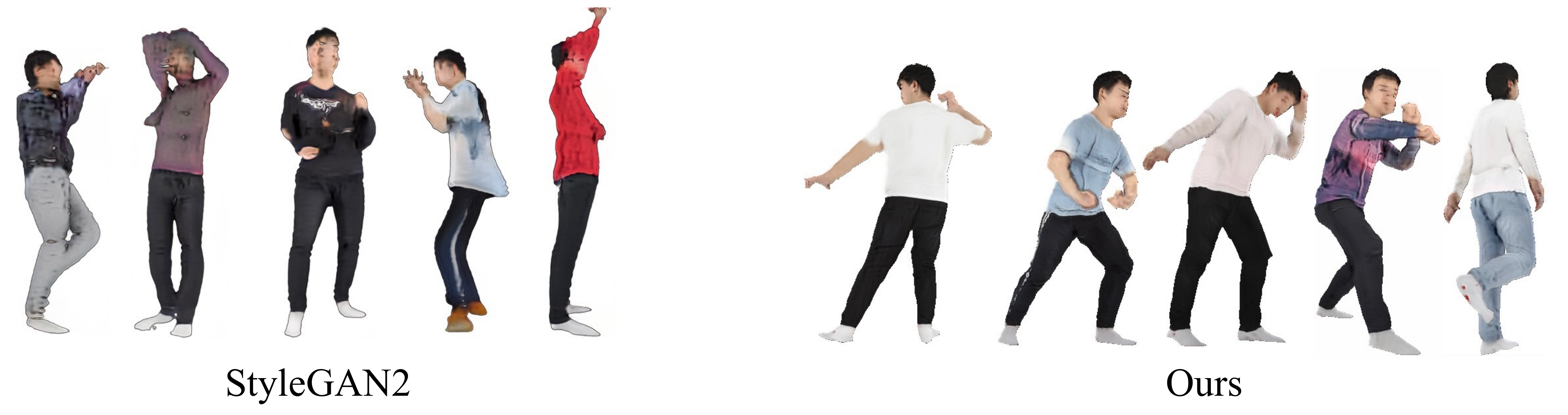}
}
\vspace{-5mm}
  \caption{Comparison of 2D StyleGAN2~\cite{karras2020analyzing} with our \nameofmethod. }
\label{fig:stylegan2_result}
\end{figure*}

We show more more generated images of the proposed method on THUman2.0~\cite{tao2021function4d}~(Fig.~\ref{fig:thu_result}) and RenderPeople~\cite{renderpeople}~(Fig.~\ref{fig:rp_result}).
We provide more  transfer learning visualization results on in-the-wild datasets:
DeepFashion~\cite{liu2016deepfashion} (Fig.~\ref{fig:df_result}) and SHHQ~\cite{fu2022stylegan} (Fig.~\ref{fig:shhq_result}).
In addition, we also make a comparison with the images generated by 2D GAN model StyleGAN2~\cite{karras2020analyzing}~(Fig.~\ref{fig:stylegan2_result}).
Please also refer to the supplementary video and 
\href{https://getavatar.github.io/}{our project page}
for more results.

\clearpage
\clearpage

\end{document}